\documentclass[10pt,twocolumn,letterpaper]{article}
\usepackage{booktabs}
\usepackage{multirow}
\usepackage{cvpr}              % To produce the CAMERA-READY version
\usepackage[ruled,linesnumbered]{algorithm2e}
\usepackage[table]{xcolor}
\usepackage{float}
\newcommand{\first}[1]{\cellcolor{red!25}{#1}}  
\newcommand{\second}[1]{\cellcolor{orange!25}{#1}} 

\definecolor{cvprblue}{rgb}{0.21,0.49,0.74}
\usepackage[pagebackref,breaklinks,colorlinks,allcolors=cvprblue]{hyperref}

\title{NimbusGS: Unified 3D Scene Reconstruction under Hybrid Weather}

\author{
Yanying Li{$^{1}$}, Jinyang Li{$^{1}$}, Shengfeng He{$^{2}$}, Yangyang Xu{$^{3}$}, Junyu Dong{$^{1}$}, Yong Du{$^{1,4,2}$}\thanks{Corresponding author (csyongdu@ouc.edu.cn).}\\
{$^{1}$}School of Computer Science and Technology, Ocean University of China, \\
{$^{2}$}Singapore Management University, {$^{3}$}Harbin Institute of Technology (Shenzhen),\\
{$^{4}$}Sanya Oceanographic Institution, Ocean University of China
}

\begin{document}

\teaser{
    \centering
    \includegraphics[width=\linewidth]{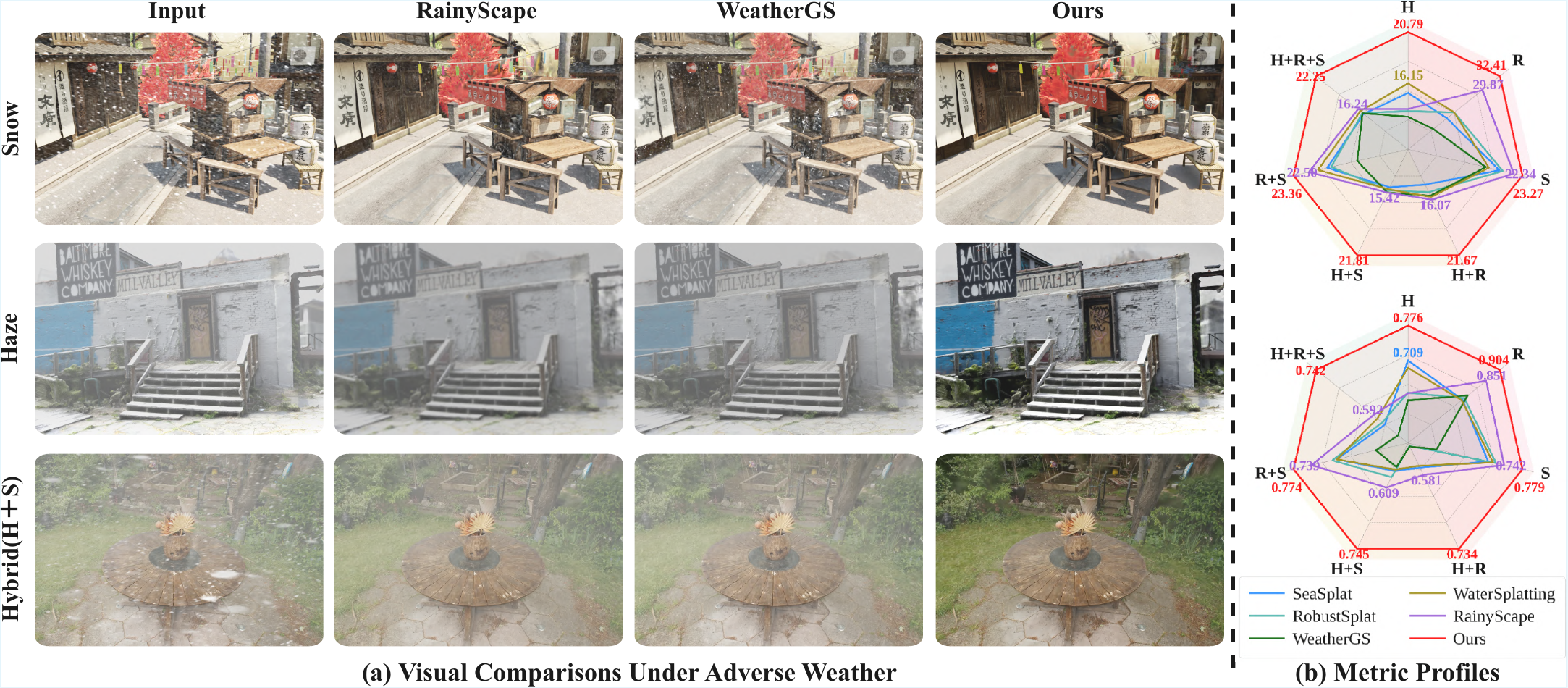}  
    \vspace{-6mm}\caption{We propose NimbusGS, a unified framework for 3D reconstruction under diverse and hybrid weather conditions. It jointly addresses continuous medium effects (haze, H), particulate degradations (snow, S; rain, R), and their mixed combinations. Panel (a) presents visual comparisons across weather types, while panel (b) summarizes metric profiles over seven single and hybrid settings.}
    \label{fig:teaser}     
}

\maketitle
\begin{abstract}
We present NimbusGS, a unified framework for reconstructing high-quality 3D scenes from degraded multi-view inputs captured under diverse and mixed adverse weather conditions. Unlike existing methods that target specific weather types, NimbusGS addresses the broader challenge of generalization by modeling the dual nature of weather: a continuous, view-consistent medium that attenuates light, and dynamic, view-dependent particles that cause scattering and occlusion. To capture this structure, we decompose degradations into a global transmission field and per-view particulate residuals. The transmission field represents static atmospheric effects shared across views, while the residuals model transient disturbances unique to each input. To enable stable geometry learning under severe visibility degradation, we introduce a geometry-guided gradient scaling mechanism that mitigates gradient imbalance during the self-supervised optimization of 3D Gaussian representations. This physically grounded formulation allows NimbusGS to disentangle complex degradations while preserving scene structure, yielding superior geometry reconstruction and outperforming task-specific methods across diverse and challenging weather conditions. Code is available at \url{https://github.com/lyy-ovo/NimbusGS}.
\end{abstract}    
\section{Introduction}\label{sec:intro}
3D scene reconstruction is a foundational task in computer vision, underpinning applications in areas such as virtual reality~\cite{zhai2025splatloc} and autonomous driving~\cite{cho2025vr,huang2024textit}. Recent advances in Neural Radiance Fields (NeRF)~\cite{mildenhall2021nerf} and 3D Gaussian Splatting (3DGS)~\cite{kerbl3Dgaussians} have significantly advanced the field, enabling photorealistic modeling of geometry and appearance from multi-view imagery. However, these methods assume clean and high-quality input, a condition rarely met in real-world environments where visual data are often corrupted by human or environmental factors. Such degradations disrupt visibility and lighting consistency, severely impairing reliable 3D reconstruction.

To address this, a growing body of work has focused on learning-based reconstruction under degraded conditions, including image blur~\cite{seiskari2024gaussian,lee2024deblurring,lu2025bard,choi2025exploiting}, underwater scenes~\cite{levy2023seathru,qiao2025restorgs,li2025watersplatting}, and low-light settings~\cite{jin2024lighting,ye2024gaussian,li2025robust,Feijoo_2025_CVPR,yan2025hvi}. While these methods achieve strong performance within their targeted domains, they are typically designed for a single degradation type or specific capture condition. In contrast, adverse weather introduces inherently more complex and entangled degradations, involving both global and localized visibility loss, which vary across time and viewpoints. These phenomena make 3D scene modeling under weather significantly more challenging than other degradation types.

From a physical perspective, weather-related degradations can be attributed to two distinct mechanisms: (i) \textit{continuous media}, such as haze and fog, which cause depth-dependent attenuation and color shifts due to scattering and absorption; and (ii) \textit{discrete particles}, including rain, snow, and sleet, which introduce dynamic, high-frequency occlusions and localized intensity changes. These two forms of degradation differ in their spatial and temporal behavior: continuous media result in smooth, view-consistent transmission effects, whereas discrete particles are transient and view-dependent. When these mechanisms co-occur, they create highly heterogeneous visual conditions that interfere with consistent geometry learning, often leading to floating artifacts and structural distortions in reconstructed scenes.

Existing approaches fall short in addressing this complexity. Some pipelines attempt to restore degraded images prior to reconstruction~\cite{li2024derainnerf,qian2025weathergs}, but this two-stage design breaks multi-view consistency and introduces rendering artifacts (\eg, WeatherGS~\cite{qian2025weathergs} in Fig.~\ref{fig:teaser}(a)). Others integrate weather modeling into the reconstruction process~\cite{chen2024dehazenerf,lyu2024rainyscape,liu2025deraings}, yet are typically restricted to specific weather types or degradation mechanisms (\eg, RainyScape~\cite{lyu2024rainyscape} in Fig.~\ref{fig:teaser}(a)). As a result, they struggle in real-world scenarios involving mixed or hybrid weather. These limitations highlight two fundamental challenges: (i) achieving generalizable 3D reconstruction across diverse weather conditions, and (ii) enabling joint modeling of continuous and discrete weather effects within a unified framework.

To tackle these challenges, we propose \textit{NimbusGS}, a coherent system for 3D scene reconstruction under both cross-weather and hybrid-weather settings. NimbusGS is \mbox{designed} to recover photorealistic geometry and appearance from multi-view inputs affected by various and overlapping weather-induced degradations. At the core of our approach is a physically informed decomposition of weather effects into two complementary components. The first is a \textit{Continuous Scattering Field}, which models global attenuation effects using a volumetric extinction field that estimates scene-wide transmittance and airlight in a view-consistent manner. The second is a \textit{Particulate Residual Layer}, which captures dynamic, view-dependent disturbances such as rain or snow through per-view residual modeling while preserving underlying scene geometry. To enable stable optimization under uneven visibility, we introduce a \textit{Geometry-Guided Gradient Scaling} mechanism that adapts gradient magnitudes based on visibility cues, improving the completeness and accuracy of distant geometry. Furthermore, a progressive optimization strategy is employed to incrementally disentangle the coupled degradation effects, leading to clearer reconstructions and improved generalization across weather conditions.

By explicitly modeling the physical nature of weather and integrating it into the reconstruction process, NimbusGS establishes a generalizable and robust solution for 3D scene modeling in real-world outdoor environments. It not only outperforms task-specific baselines across individual weather types, but also sets a new benchmark in complex hybrid-weather scenarios (see Fig.~\ref{fig:teaser}(b)).

Our main contributions are fourfold:
\begin{itemize}
    \item We propose NimbusGS, a unified framework for 3D reconstruction under diverse and hybrid weather, enabling physically grounded and generalizable modeling across complex degradations.
    
    \item We design Continuous Scattering Modeling and Particulate Layer Modeling to separately capture volumetric transmission and dynamic particulate residuals, enabling interpretable weather decomposition.
    
    \item We introduce Geometry-Guided Gradient Scaling to regulate geometry updates under uneven visibility, reducing artifacts and improving completeness in occluded regions.
    
    \item NimbusGS achieves state-of-the-art performance in haze, rain, snow, and mixed-weather settings, without requiring paired data or large-scale pretraining.
\end{itemize}

\section{Related Work}\label{sec:formatting}

\noindent\textbf{Radiance Fields.}
Neural Radiance Fields (NeRF)~\cite{mildenhall2021nerf} and its variants~\cite{garbin2021fastnerf,muller2022instant,reiser2021kilonerf,sun2022direct} learn continuous volumetric representations for novel view synthesis and 3D reconstruction. 3D Gaussian Splatting (3DGS)~\cite{kerbl3Dgaussians} improves rendering efficiency by replacing volume integration with forward $\alpha$-blending of Gaussian primitives. Recent extensions further enhance radiance-field robustness under sparse views~\cite{paliwal2024coherentgs,zhang2024cor,charatan2024pixelsplat,zheng2025nexusgs} and reduce artifacts~\cite{lin2025hybridgs,sabour2025spotlesssplats}, improving view consistency and geometric coherence. Nevertheless, radiance-field representations remain sensitive to degraded visibility, often leading to unstable geometry. NimbusGS addresses this by introducing geometry-guided gradient scaling to stabilize optimization under adverse weather.

\noindent\textbf{Image and Video Restoration.}
Image restoration removes degradations such as blur, noise, and weather artifacts. Recent works leverage depth and frequency priors~\cite{Zhang_2024_CVPR,zhou2024seeing,wang2024depth,wang2024selfpromer,chen2024teaching} or diffusion-based models~\cite{Liu_2024_CVPR,xu2024boosting,ye2024learning,chen2025unirestore} for condition-aware enhancement. Unified frameworks~\cite{conde2024instructir,ai2024multimodal,liu2024diff,luo2024controlling} integrate multiple restoration tasks. For videos, motion alignment and temporal modeling~\cite{kim2024frequency,zhang2024avid,zhang2024blur,youk2024fma} improve coherence. Yet, these methods lack 3D awareness and often fail under multi-view degradation. NimbusGS instead operates in 3D space, producing view-consistent results without paired supervision or large-scale pretraining.

\noindent\textbf{3D Reconstruction under Adverse Weather.}
Weather introduces complex degradations that impair geometry and appearance modeling. Some methods integrate physical priors, such as DehazeNeRF~\cite{chen2024dehazenerf}, SeaThru-NeRF~\cite{levy2023seathru}, and ScatterNeRF~\cite{hu2025scattersplatting}, to handle haze and scattering. Others adapt 3DGS to specific weather types: DeRainGS~\cite{liu2025deraings} uses paired supervision for rain; WeatherGS~\cite{qian2025weathergs} separates particles from lens artifacts; DehazeGS~\cite{yu2025dehazegs} estimates transmission for clean rendering. These methods often target a single degradation and rely on 2D priors, limiting generalization and geometric fidelity. NimbusGS jointly models continuous scattering and dynamic particles within a unified 3D framework, enabling consistent reconstruction across diverse and hybrid-weather conditions.

\section{Preliminaries}
\noindent\textbf{Gaussian Splatting.}
3D Gaussian Splatting (3DGS)~\cite{kerbl3Dgaussians} explicitly represents a scene as a set of anisotropic 3D Gaussians parameterized by their centers, covariances, opacities, and colors $(\mu,\,\Sigma,\,o,\,c)$.
Gaussians along each camera ray are sorted from near to far and accumulated via $\alpha$-blending:
\begin{equation}
C = \sum_i c_i\,\alpha_i \prod_{j=1}^{i-1}(1-\alpha_j),
\end{equation}
where $C$ denotes the final pixel color, $c_i$ denotes the color of the $i$-th Gaussian, and $\alpha_i \in (0,1)$ denotes its per-Gaussian opacity at the pixel.
The cumulative product term represents the accumulated transmittance before the $i$-th Gaussian.

For optimization, 3DGS typically adopts the L1 and D-SSIM losses between rendered and observed images during training, enabling photorealistic reconstruction of clean scenes.
However, when the input images are degraded by adverse weather (\eg, haze, rain, or snow), the Gaussian parameters inevitably absorb degradation cues into the geometry and appearance fields, leading to inconsistent radiance and floating artifacts across views.

\noindent\textbf{Radiative Transfer Equation.}
Image formation in participating media (\eg, haze or fog) follows the Radiative Transfer Equation (RTE)~\cite{chandrasekhar2013radiative}, modeling light attenuation and in-scattering. Radiance transport along a viewing ray is parameterized as
\begin{equation}
\mathbf r(s) = \mathbf o + s\,\mathbf d,\quad s \in [t_n, t_f],
\label{eq:ray}
\end{equation}
where $\mathbf o$ and $\mathbf d$ denote the camera origin and ray direction, and $s$ represents distance from near to far bounds $t_n,t_f$. At position $\mathbf x$, the medium's optical property is described by the extinction coefficient
\begin{equation}
\sigma_t(\mathbf x) = \sigma_a(\mathbf x) + \sigma_s(\mathbf x),
\label{eq:extinction}
\end{equation}
where $\sigma_a$ and $\sigma_s$ denote the absorption and scattering coefficients, respectively. According to the Beer–Lambert law, the transmittance from $t_n$ to $s$ is
\begin{equation}
\tilde{T}(s) = \exp\!\Big(-\!\int_{t_n}^{s}\sigma_t\!\big(\mathbf r(u)\big)\,du\Big),
\label{eq:transmittance}
\end{equation}
representing the fraction of radiance that remains unabsorbed and unscattered along the optical path.

With these definitions, the RTE is expressed as
\begin{equation}
\begin{aligned}
L(\mathbf o,\omega)
&= \underbrace{\tilde{T}(t_0)\,L_{\text{surf}}\!\big(\mathbf r(t_0),\omega\big)}_{\text{surface attenuation term}} \\
&\quad + \underbrace{\int_{t_n}^{t_0}\! \tilde{T}(s)\,\sigma_s\!\big(\mathbf r(s)\big)\,A\!\big(\mathbf r(s),\omega\big)\,ds}_{\text{airlight scattering term}}.
\end{aligned}
\label{eq:rte}
\end{equation}
where $L(\mathbf o,\omega)$ is the radiance received at the camera position $\mathbf o$ in direction $\omega$, and $t_0$ denotes the first surface intersection on the ray. $L_{\text{surf}}$ represents the surface-emitted radiance without participating media, while $A(\mathbf x,\omega)$ denotes the spatially varying airlight color determined by position and viewing direction.

\begin{figure*}[t]
	\centering
	\includegraphics[width=0.98\linewidth]{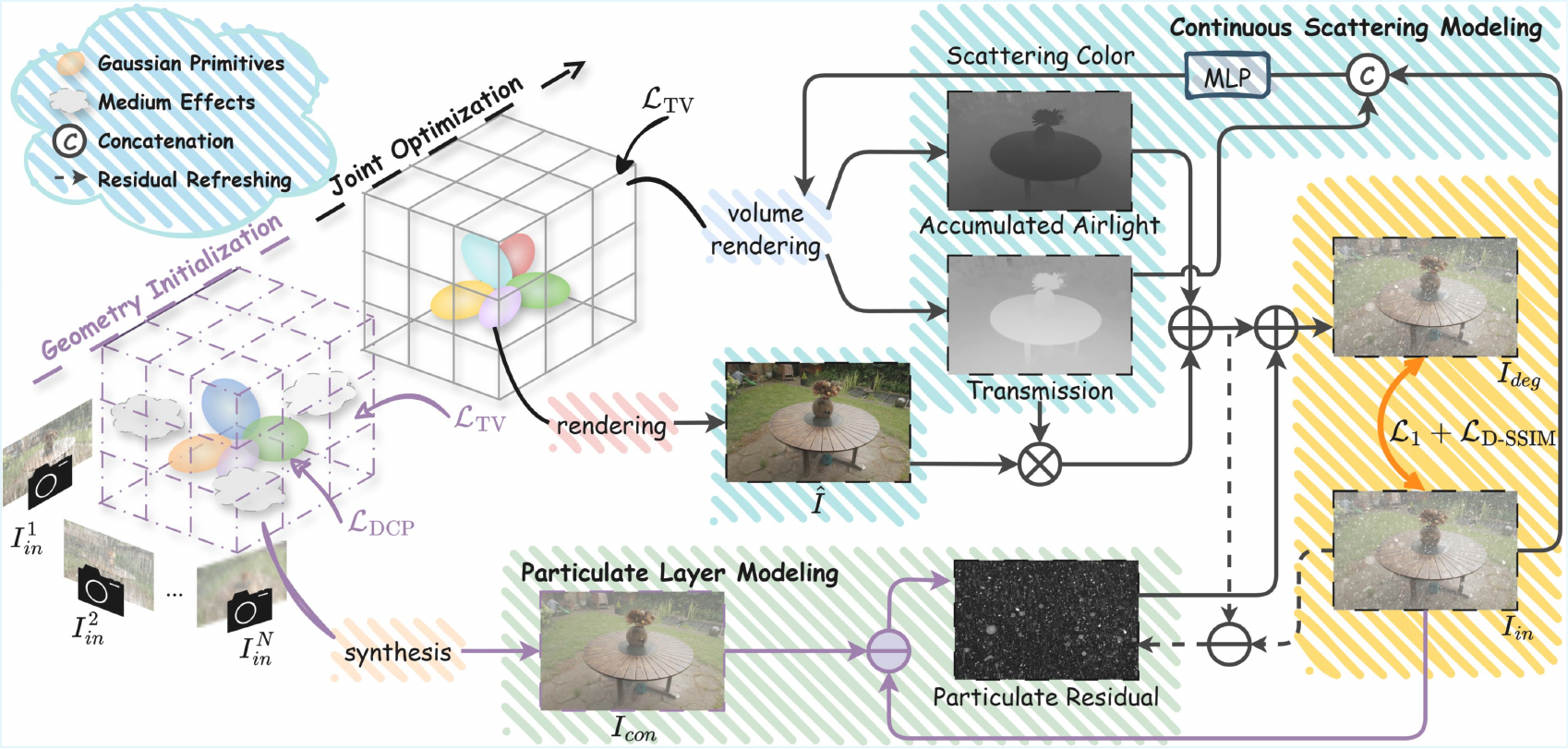}
	\vspace{-2mm}\caption{Overview of NimbusGS. Starting from a geometry initialization, transient particle effects are separated as per-view residuals. CSM estimates an extinction field from which transmission and airlight are derived, blended with the scene rendering and residuals to reproduce the degradations. This self-supervised process guides the Gaussian representation toward a clean and consistent reconstruction.}\vspace{-6mm}
	\label{fig:pipeline}
\end{figure*}

\section{Method}
\noindent\textbf{Problem Formulation.}
NimbusGS models two weather-induced degradations within a unified differentiable rendering framework: a view-consistent continuous medium that attenuates light, and a view-dependent particulate layer that introduces transient occlusions across views. In particular, we formulate the degraded rendering $I_{\text{deg}}$ as:
\begin{equation}
I_{\text{deg}} = \hat{I} \cdot T + P + R,
\label{eq:1}
\end{equation} 
where $\hat I$ denotes the rendered image from the learned clean radiance field, $T$ is the transmission map, $P$ is the airlight scattering term, and $R$ is the particulate layer capturing dynamic weather artifacts. The first two terms represent the continuous-medium degradation, while the last term accounts for the view-dependent particulate effects.

An overview of our pipeline is illustrated in Fig.~\ref{fig:pipeline}. During training, NimbusGS operates in a self-supervised manner using only $N$ degraded input views $\{I_{in}^i\}_{i=1}^{N}$.

\noindent\textbf{Continuous Scattering Modeling (CSM).} 
We represent the continuous medium as a voxelized extinction field $\beta(\mathbf{x})$, approximating the total extinction $\sigma_t(\mathbf{x})$ and capturing spatially varying absorption and scattering. To ensure full volumetric coverage of the Gaussian field, the grid is initialized within an expanded bounding box enclosing all Gaussians, preventing edge scattering loss during ray integration. The continuous-medium degradation is then expressed as
\begin{equation}
I_{\text{con}} = \hat I \cdot T + P,
\label{eq:Ideg}
\end{equation}
from which the transmission $T$ and airlight term $P$ are derived through our volumetric formulation.

Specifically, we first uniformly sample $K$ points within the interval $[t_n, t_0]$ along each camera ray and compute the transmittance as a discrete form of Eq.~\eqref{eq:transmittance}:
\begin{equation}
T = \exp\!\Big(-\sum_{j=1}^{K}\beta\!\big(\mathbf r(s_j)\big)\,\Delta s_j\Big),
\label{eq:disc-trans}
\end{equation}
where $s_j$ denotes the \(j\)-th sampling position, and $\Delta s_j = s_j - s_{j-1}$ is the distance between adjacent samples.

We further denote the cumulative transmittance up to the \(i\)-th sampling point as 
\begin{equation}
T_i = \exp\!\Big(-\sum_{j=1}^{i-1}\beta(\mathbf r(s_j))\,\Delta s_j\Big),
\label{eq:Ti}
\end{equation}
which represents the fraction of radiance that remains unattenuated before reaching the \(i\)-th slice. Under the single-scattering assumption, the airlight term $P$ is discretized from Eq.~(\ref{eq:rte}) using $\alpha$-blending:
\begin{equation}
P = \sum_{i=1}^{K} T_i\,\Big(1 - \exp\!\big(-\beta(\mathbf r(s_i))\,\Delta s_i\big)\Big)\,A,
\label{eq:C}
\end{equation}
where the scattering color $A$ is predicted by an MLP conditioned on the degraded image $I_{in}$ and transmission $T$. 

Note that although the proposed CSM adopts the single-scattering assumption for computational simplicity, its spatially varying extinction field and learned airlight component still provide an effective statistical approximation of multiple-scattering effects. By linking transmission and airlight through the shared extinction representation, our method also mitigates the ambiguity that arises when these terms are estimated independently, ensuring consistent decomposition across views. This formulation thus retains the tractability of single-scattering rendering while improving fidelity in realistic, non-uniform scattering environments.

\noindent\textbf{Particulate Layer Modeling (PLM).}
While the continuous medium models spatially smooth attenuation, transient weather particles cause view-dependent occlusions. To explicitly separate these dynamic effects from the static scene, we introduce a particulate layer modeling scheme.

Specifically, we decompose the degraded rendering $I_{\text{deg}}$ into a continuous-medium rendering $I_{\text{con}}$ and a view-dependent particulate layer $R$:
\begin{equation}
I_{\text{deg}} = I_{\text{con}} + R.
\end{equation}
During early training of 3DGS, the Gaussians, typically initialized with large scales and low opacities, remain in this state, biasing the model toward view-consistent low-frequency geometry while causing high-frequency dynamic signals to be excluded from the learned representation.

We leverage this property through a geometry initialization stage, during which a reduced learning rate is applied to the Gaussian scale parameters. This stage stabilizes the underlying geometric structure while preventing transient particles from being absorbed into it. Although the involved components are not yet fully disentangled due to incomplete learning, $I_{\text{con}}$ still reflects the static portion of the scene without dynamic interference, allowing the particulate layer to be isolated by subtraction:
\begin{equation}
R = \mathrm{ReLU}(I_{\text{in}} - I_{\text{con}}),
\end{equation}
where the ReLU operator suppresses negative differences and preserves additive residuals induced by transient particles. The resulting $R$ serves as an initial particulate residual and is periodically updated during later optimization: as the model parameters are refined, a new $I_{\text{con}}$ is rendered and a refreshed residual is re-extracted. This process keeps the particulate layer view-dependent and decoupled from the evolving static geometry.

\noindent\textbf{Geometry-Guided Gradient Scaling (GGS).}
Both continuous and particulate degradations attenuate distant radiance, reducing pixel contrast and suppressing reconstruction gradients in far-depth regions. In standard 3DGS, densification is triggered by an accumulated-gradient threshold. Consequently, distant geometry often fails to meet this threshold and remains under-densified, while large Gaussians with wide screen projections blur fine details by covering excessive image regions.

To address this issue, we introduce a Geometry-Guided Gradient Scaling mechanism that adaptively rescales per-Gaussian gradients based on three complementary factors: depth, projected radius, and reconstruction error. This adjustment amplifies gradients for distant or oversized Gaussians, enabling more balanced densification and sharper geometric reconstruction.

For each Gaussian $i$ with depth $d_i$ and projected coordinate $(u_i,v_i)$, we first normalize depth for scale consistency:
\begin{equation}
d_{\text{norm}}^i = \frac{d_i - d_{\min}}{d_{\max} - d_{\min}} \in [0,1].
\end{equation}
Next, we obtain the per-Gaussian error $e_i$ by bilinear sampling at $(u_i,v_i)$:
\begin{equation}
e_i = \mathrm{bilinear}\!\left(\|I_{\text{deg}} - I_{\text{in}}\|_1, (u_i,v_i)\right).
\end{equation}
Since the sampled errors may include outliers, we further apply a median-based normalization for robustness:
\begin{equation}
e_{\text{norm}}^i = 
\frac{e_i - \mathrm{median}(\{e_i\})}
{\mathrm{median}(|e_i - \mathrm{median}(\{e_i\})|)}.
\end{equation}
The normalized error is then passed through a sigmoid function $\sigma(\cdot)$ to obtain a bounded gate for gradient computation.

Finally, the overall scaling weight is computed as
\begin{equation}
w_i = d_{\text{norm}}^i \cdot 
\left(\frac{r_i}{r_0}\right) \cdot 
\sigma(e_{\text{norm}}^i),
\end{equation}
where $r_i$ is the Gaussian’s projected radius and $r_0$ is a reference radius constant. Then, we mean-normalize the weights and rescale gradients:
\begin{equation}
\overline{w} = \frac{1}{N}\sum_i w_i, \qquad
\nabla \theta_i \leftarrow 
\mathrm{sg}\!\left(\frac{w_i}{\overline{w}}\right)\! \cdot \nabla \theta_i,
\end{equation}
where $N$ is the total number of Gaussians, $\theta_i$ represents the parameters of the $i$-th Gaussian, and $\mathrm{sg}(\cdot)$ denotes the stop-gradient operator. This mean normalization keeps the average scaling factor at 1, thus maintaining global gradient magnitude while redistributing relative update strengths among Gaussians. As a result, distant structures receive stronger gradients for improved densification, and large-radius Gaussians are regularized, yielding clearer and more consistent geometry under adverse weather degradations.

\noindent\textbf{Training Strategy and Objective.}  
NimbusGS uses a two-stage training scheme: the first separates particulate effects from static geometry to initialize the residual, and the second jointly refines the Gaussian representations and the degradation components.

In the first stage, we use a photometric loss that mixes pixel-wise and structural similarity between $I_{\text{in}}$ and $I_{\text{con}}$, together with two regularization terms $\mathcal L_{\text{DCP}}$ and $\mathcal L_{\text{TV}}$. The total objective $\mathcal L_{\text{ini}}$ is given by
\begin{equation}
\begin{aligned}\small
\mathcal L_{\text{ini}} =&(1-\lambda_{\text{r}})\|I_{\text{in}}-I_{\text{con}}\|_{1} \\
&+\lambda_{\text{r}}(1-\mathrm{SSIM}(I_{\text{in}},I_{\text{con}}))
+\mathcal L_{\text{DCP}}+\mathcal L_{\text{TV}},
\end{aligned}
\label{eq:render1}
\end{equation}
where $\lambda_{\text{r}}$ is the balancing factor. After this brief warm-up stage, once the geometry stabilizes, the particulate layer $R$ is initialized from the residual between $I_{\text{in}}$ and $I_{\text{con}}$.

We next detail regularization terms. Following the dark channel prior~\cite{he2010single}, we regularize the rendered clean image $\hat{I}$ to suppress excessive scattering in the continuous medium:
\begin{equation}
\mathcal L_{\text{DCP}}=\|\mathrm{DCP}(\hat{I})\|_{1}.
\label{eq:dcp}
\end{equation}

\begin{table*}[!t]
\centering
\setlength{\tabcolsep}{3.6pt}
\resizebox{\textwidth}{!}{
\begin{tabular}{l ccc ccc ccc ccc ccc}
\toprule
\multirow{2}{*}{Method} &
\multicolumn{3}{c}{Bicycle} &
\multicolumn{3}{c}{Garden} &
\multicolumn{3}{c}{Factory} &
\multicolumn{3}{c}{Tanabata} &
\multicolumn{3}{c}{Average} \\
\cmidrule(lr){2-4}\cmidrule(lr){5-7}\cmidrule(lr){8-10}\cmidrule(lr){11-13}\cmidrule(lr){14-16}
& PSNR$\uparrow$ & SSIM$\uparrow$ & LPIPS$\downarrow$
& PSNR$\uparrow$ & SSIM$\uparrow$ & LPIPS$\downarrow$
& PSNR$\uparrow$ & SSIM$\uparrow$ & LPIPS$\downarrow$
& PSNR$\uparrow$ & SSIM$\uparrow$ & LPIPS$\downarrow$
& PSNR$\uparrow$ & SSIM$\uparrow$ & LPIPS$\downarrow$ \\
\midrule
3DGS~\cite{kerbl3Dgaussians} 
& 17.19 & 0.578 & 0.319 
& 13.33 & 0.522 & 0.344 
& 15.51 & 0.687 & 0.206 
& 12.25 & 0.565 & 0.262 
& 14.57 & 0.588 & 0.283 \\
SeaSplat~\cite{yang2025seasplat} 
& \second{18.28} & 0.588 & 0.312 
& 15.07 & \second{0.676} & 0.326 
& 12.50 & 0.792 & 0.190 
& \second{16.44} & \second{0.779} & \second{0.185} 
& 15.57 & \second{0.709} & 0.253 \\
WaterSplatting~\cite{li2025watersplatting} 
& 18.19 & \second{0.593} & \second{0.301} 
& \second{15.25} & 0.655 & \second{0.318} 
& \second{16.15} & \second{0.814} & \second{0.182} 
& 15.02 & 0.716 & 0.203 
& \second{16.15} & 0.695 & \second{0.251} \\
\textbf{Ours}
& \first{21.84} & \first{0.677} & \first{0.249}
& \first{21.04} & \first{0.727} & \first{0.274}
& \first{21.16} & \first{0.899} & \first{0.104}
& \first{19.12} & \first{0.802} & \first{0.131}
& \first{20.79} & \first{0.776} & \first{0.190} \\
\bottomrule
\end{tabular}}
\vspace{-2mm}
\captionof{table}{Quantitative comparisons on hazy scenes. The \colorbox{red!25}{best} and \colorbox{orange!25}{second-best} scores are color-encoded for clarity.}
\vspace{-2mm}
\label{tab:dehaze}
\end{table*}

\begin{figure*}[!t]
  \centering
  \begin{minipage}[t]{0.195\linewidth}
  \centering  
    \includegraphics[width=\linewidth]{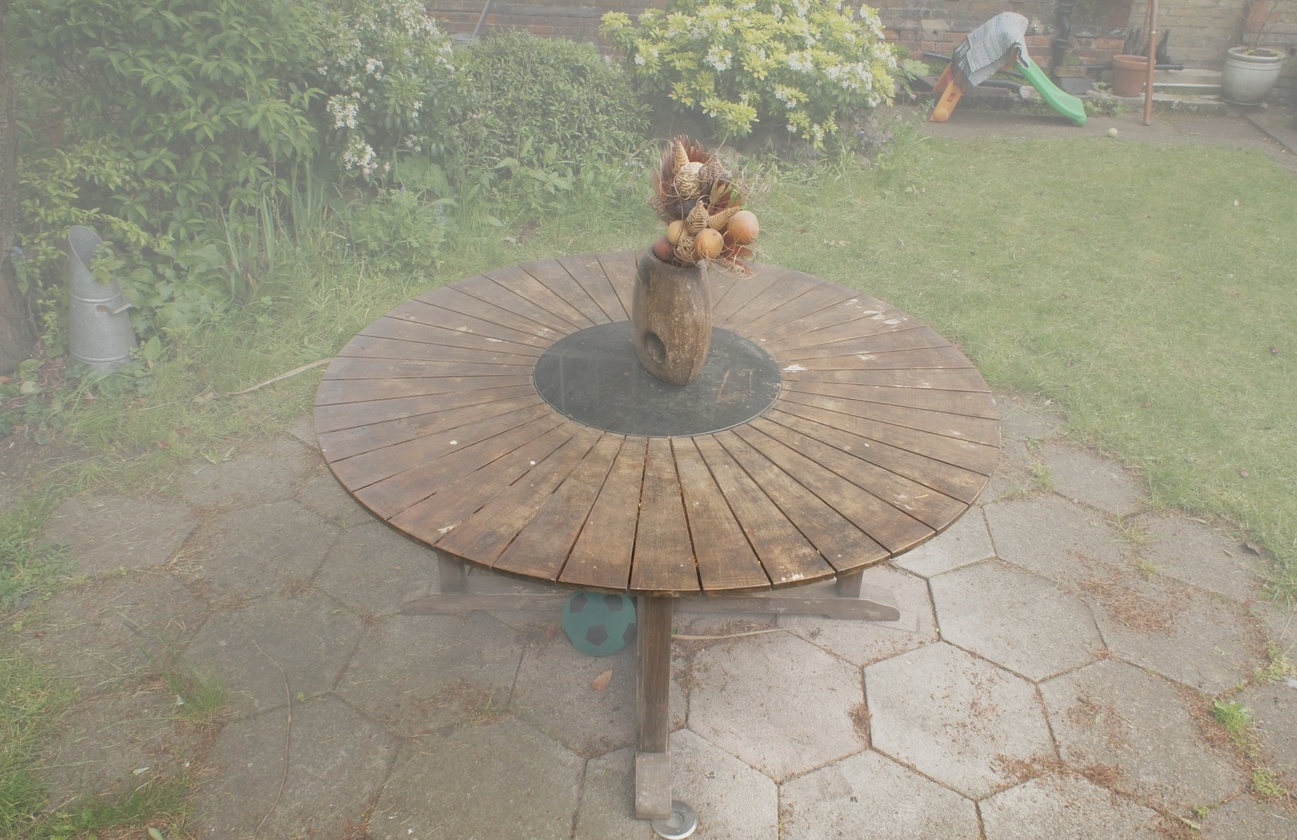}
    {\small Input}
  \end{minipage}\hfill
  \begin{minipage}[t]{0.195\linewidth}
  \centering   
    \includegraphics[width=\linewidth]{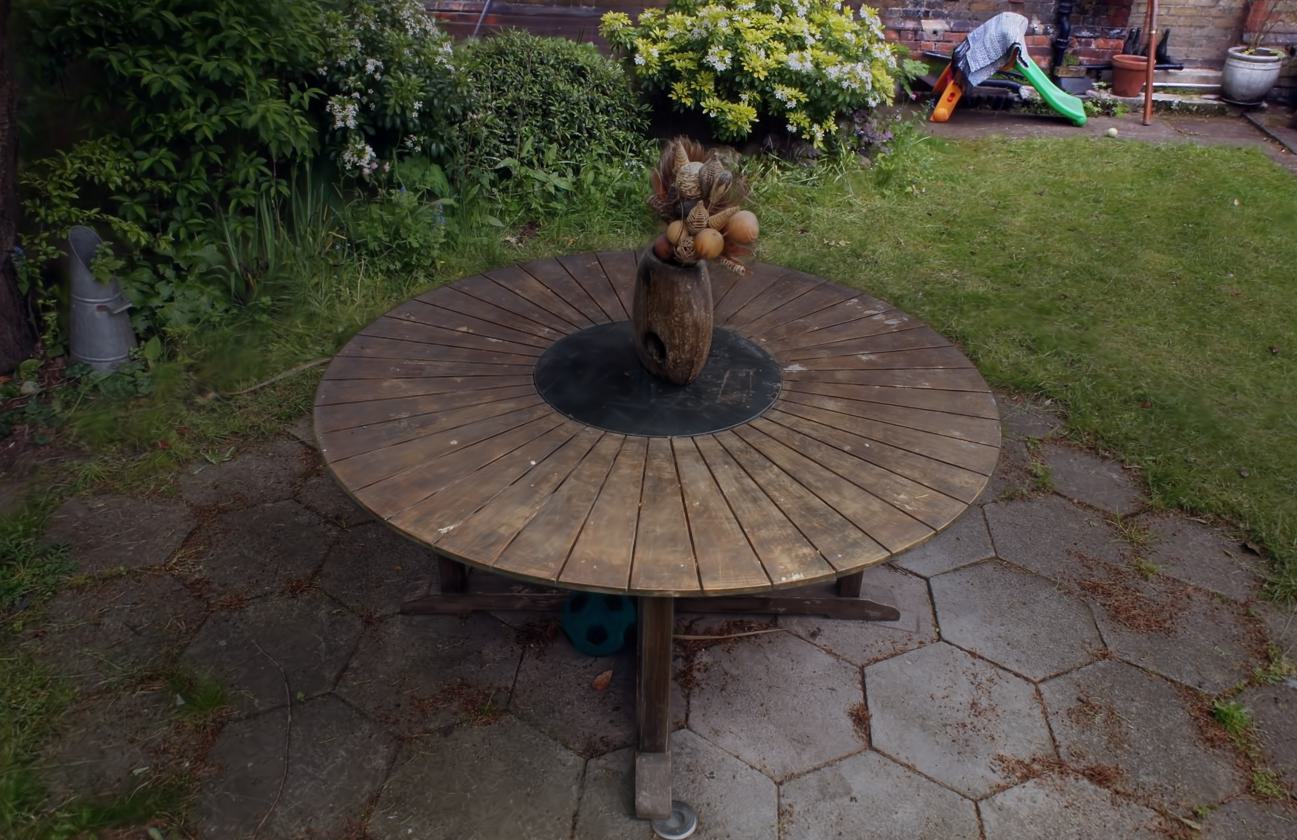}
    {\small SeaSplat~\cite{yang2025seasplat}}
  \end{minipage}\hfill
  \begin{minipage}[t]{0.195\linewidth}
  \centering   
    \includegraphics[width=\linewidth]{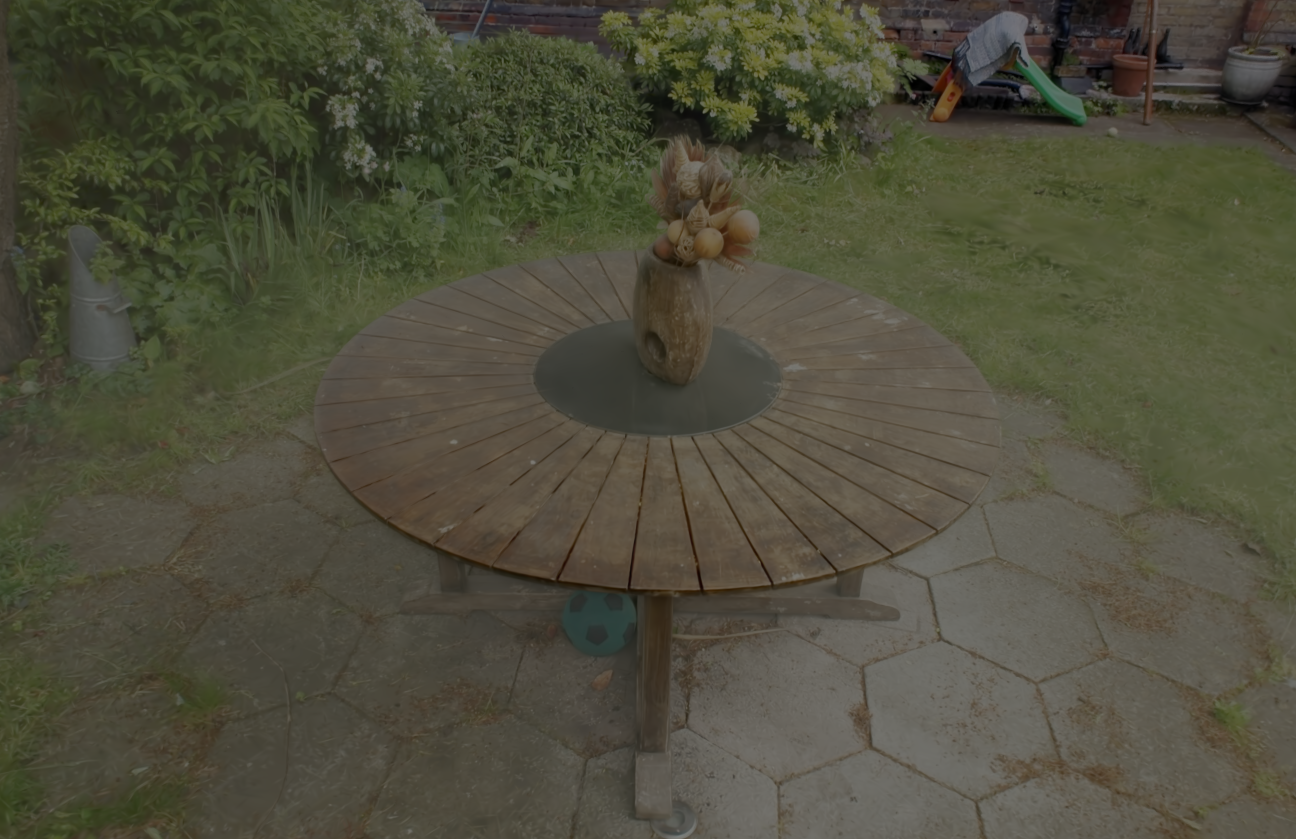}
    {\small WaterSplatting~\cite{li2025watersplatting}}
  \end{minipage}\hfill
  \begin{minipage}[t]{0.195\linewidth}
  \centering  
    \includegraphics[width=\linewidth]{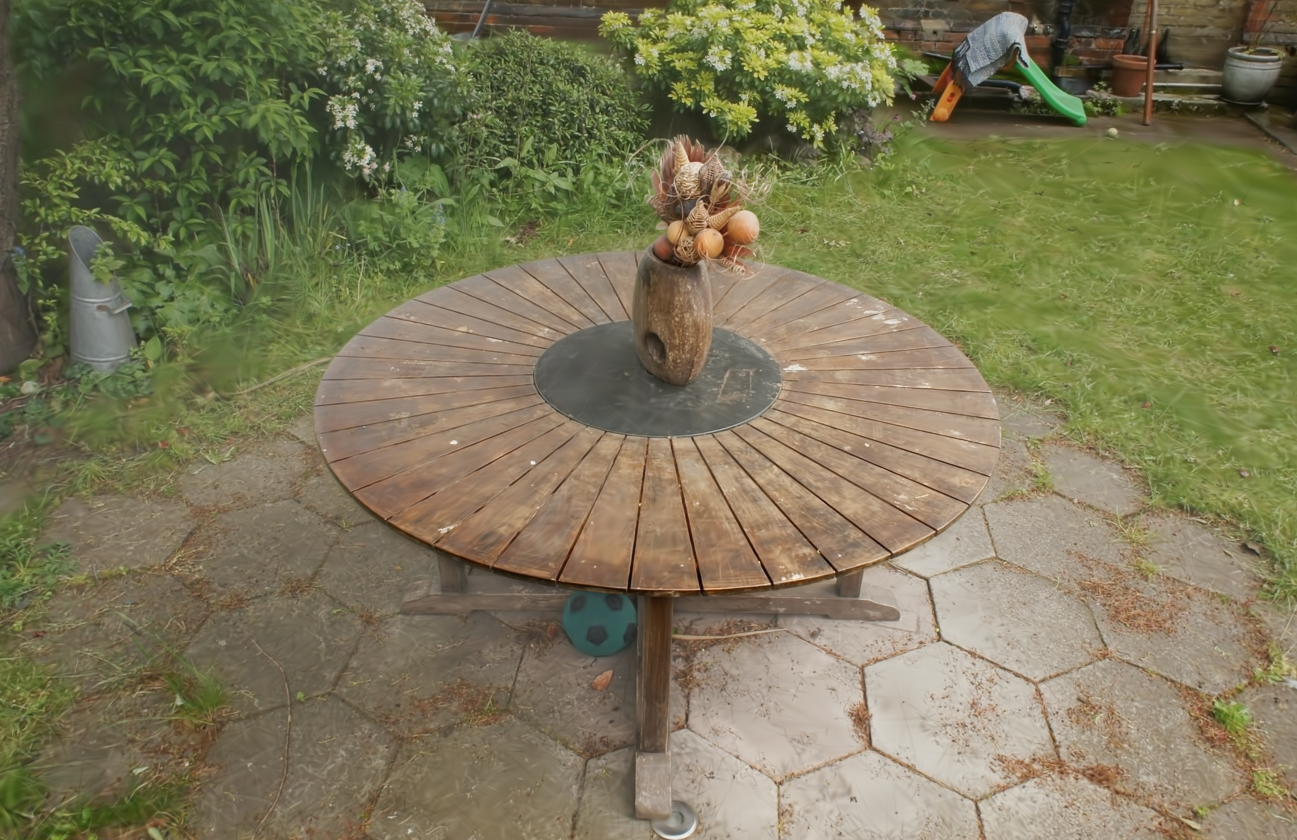}
    {\small Ours}
  \end{minipage}\hfill
  \begin{minipage}[t]{0.195\linewidth}
  \centering   
    \includegraphics[width=\linewidth]{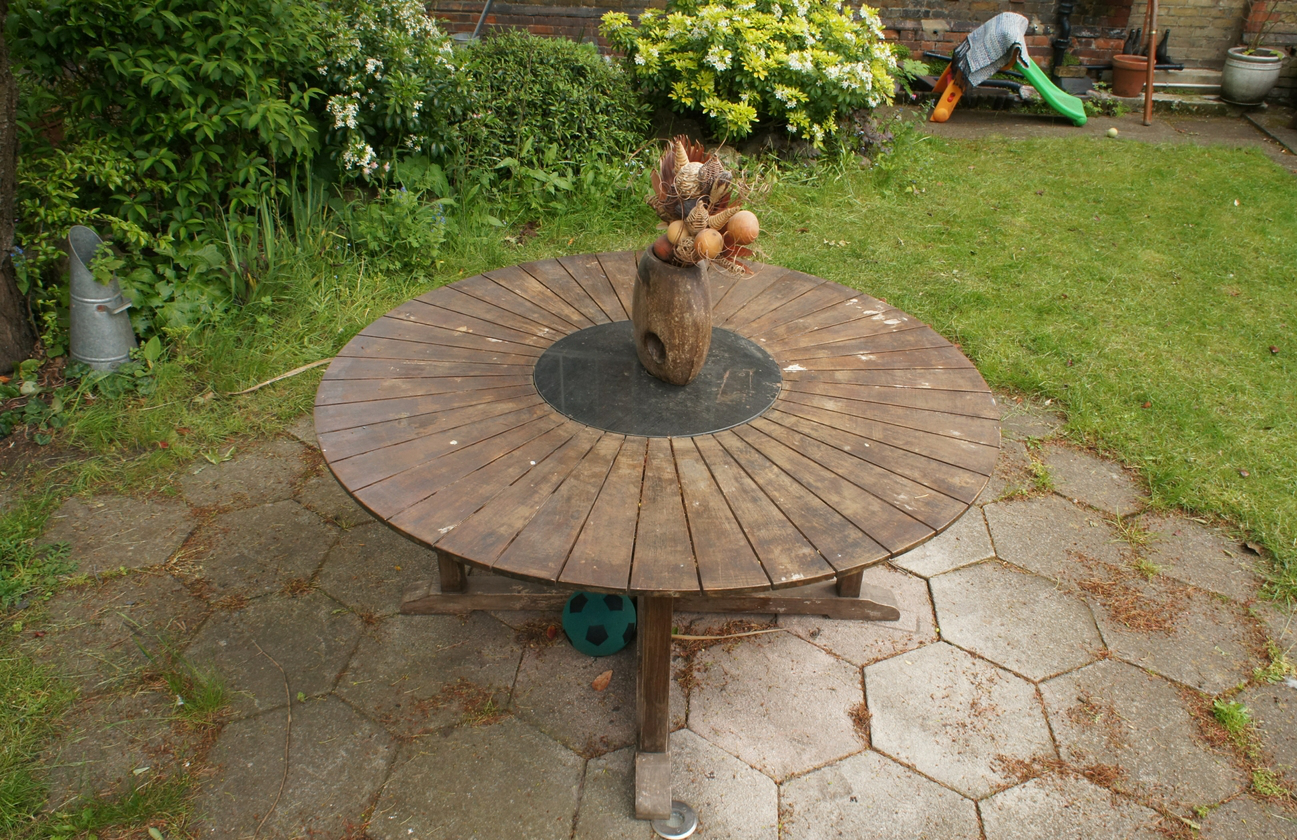}
    {\small Ground Truth}
  \end{minipage}
  \vspace{-2mm}
  \caption{Qualitative results on hazy scenes. Best viewed zoomed in.}\vspace{-4mm}
  \label{fig:fog}  
\end{figure*}

To smooth the extinction field while preserving structural boundaries, we apply total-variation regularization: 
\begin{equation}
\mathcal L_{\text{TV}}=\|\nabla\beta\|_1.
\label{eq:tv}
\end{equation}

In the second stage, we jointly optimize the Gaussian primitives and the continuous medium while periodically refreshing the particulate layer as training proceeds. The photometric loss is now computed between the reconstructed degraded image $I_\text{deg}$ and the input observation $I_\text{in}$. The dark channel prior is removed, as it relies on natural image statistics assuming locally low-intensity pixels in clear scenes, which may bias the reconstruction toward under-exposed solutions during later optimization. The objective $\mathcal L$ is defined as
\begin{equation}
\begin{aligned}\small
\mathcal L =&(1-\lambda_{\text{r}})\|I_{\text{in}}-I_{\text{deg}}\|_{1}+ \lambda_{\text{r}}\!\left(1-\mathrm{SSIM}(I_{\text{in}},I_{\text{deg}})\right)\\
&+ \mathcal L_{\text{TV}}.
\end{aligned}
\label{eq:total}
\end{equation}

\begin{table*}[!t]
\centering
\setlength{\tabcolsep}{3.6pt}
\resizebox{\textwidth}{!}{
\begin{tabular}{l ccc ccc ccc ccc ccc}
\toprule
\multirow{2}{*}{Method} &
\multicolumn{3}{c}{Crossroad} &
\multicolumn{3}{c}{Square} &
\multicolumn{3}{c}{Sailboat} &
\multicolumn{3}{c}{Yard} &
\multicolumn{3}{c}{Average} \\
\cmidrule(lr){2-4}\cmidrule(lr){5-7}\cmidrule(lr){8-10}\cmidrule(lr){11-13}\cmidrule(lr){14-16}
& PSNR$\uparrow$ & SSIM$\uparrow$& LPIPS$\downarrow$
& PSNR$\uparrow$ & SSIM$\uparrow$& LPIPS$\downarrow$
& PSNR$\uparrow$ & SSIM$\uparrow$& LPIPS$\downarrow$
& PSNR$\uparrow$ & SSIM$\uparrow$& LPIPS$\downarrow$
& PSNR$\uparrow$ & SSIM$\uparrow$& LPIPS$\downarrow$ \\
\midrule
DerainNeRF~\cite{li2024derainnerf}
& 23.97 & 0.623 & 0.381
& 22.71 & 0.737 & 0.370
& 23.73 & 0.805 & 0.192
& 22.86 & 0.765 & 0.253
& 23.31 & 0.732 & 0.299 \\
RobustSplat~\cite{2025RobustSplat}
& 27.45 & 0.729 & 0.308
& 26.35 & 0.793 & 0.319
& 26.17 & 0.751 & 0.340
& 27.57 & 0.796 & 0.297
& 26.88 & 0.767 & 0.316 \\
WeatherGS~\cite{qian2025weathergs}
& 25.32 & 0.653 & 0.345
& 24.87 & 0.807 & 0.309
& 26.52 & 0.839 & 0.195
& 25.57 & \second{0.824} & 0.250
& 25.57 & 0.780 & 0.274 \\
RainyScape~\cite{lyu2024rainyscape}
& \second{31.01} & \second{0.851} & \first{0.154}
& \second{29.44} & \second{0.883} & \second{0.152}
& \second{29.62} & \second{0.851} & \second{0.147}
& \second{29.43} & 0.820 & \second{0.166}
& \second{29.87} & \second{0.851} & \second{0.154} \\
\textbf{Ours}
& \first{32.15} & \first{0.893} & \second{0.189}
& \first{32.13} & \first{0.914} & \first{0.145}
& \first{33.19} & \first{0.921} & \first{0.109}
& \first{32.18} & \first{0.889} & \first{0.116}
& \first{32.41} & \first{0.904} & \first{0.139} \\
\bottomrule
\end{tabular}}
\vspace{-2mm}
\captionof{table}{Quantitative comparisons on rainy scenes. The \colorbox{red!25}{best} and \colorbox{orange!25}{second-best} scores are color-encoded for clarity.}
\vspace{-2mm}\label{tab:derain}
\end{table*}

\section{Experiments}
\begingroup
\newcommand{\colw}{0.136\linewidth}  
\newcommand{\colgap}{0.1pt}          
\begin{figure*}[!t]
  \centering 
  \begin{minipage}[t]{\colw}
  \centering
    \includegraphics[width=\linewidth]{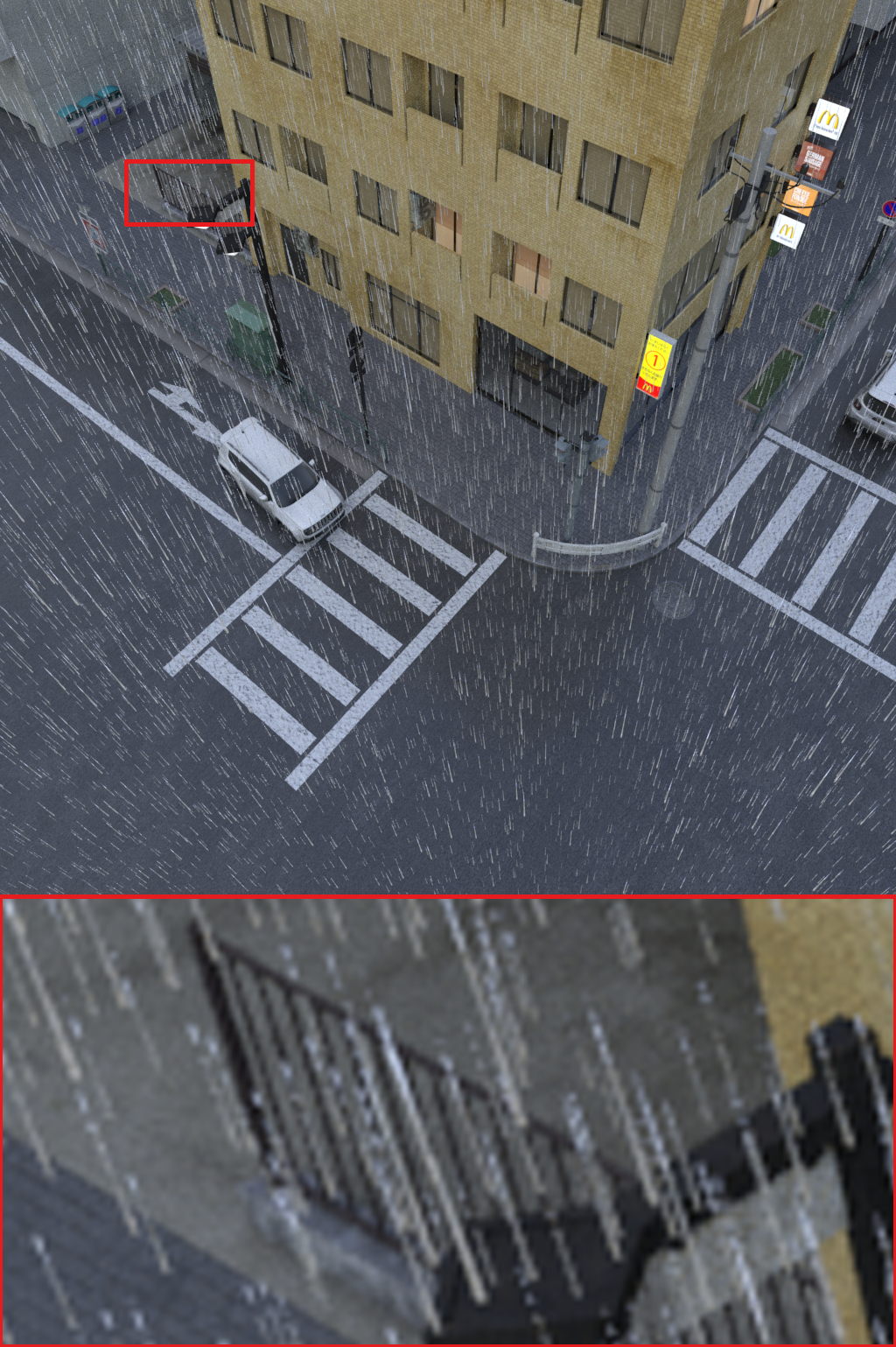}  
    {\small Input}
  \end{minipage}\hspace{\colgap}
  \begin{minipage}[t]{\colw}
  \centering
    \includegraphics[width=\linewidth]{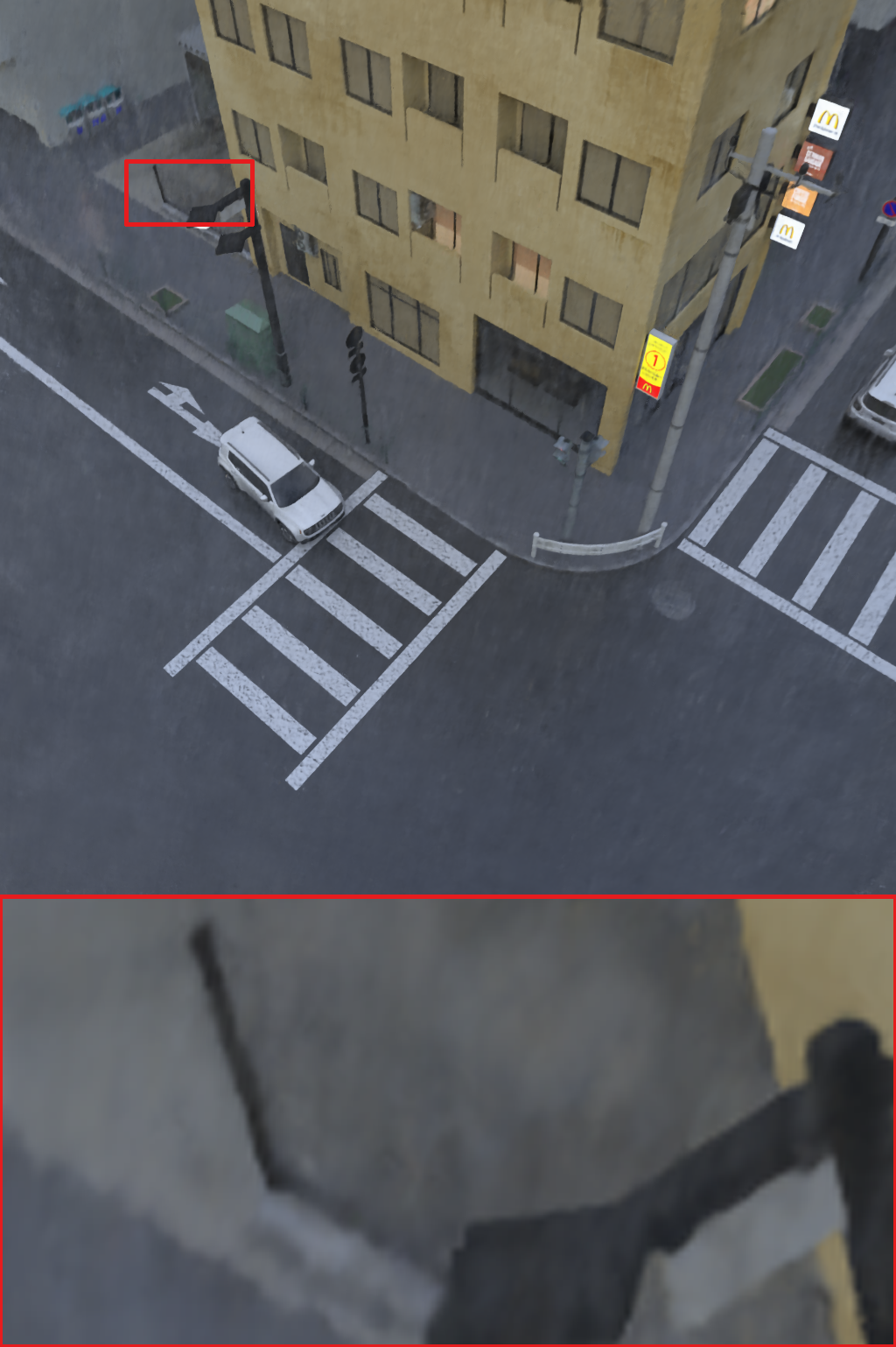}   
    {\small DerainNeRF~\cite{li2024derainnerf}}
  \end{minipage}\hspace{\colgap}
  \begin{minipage}[t]{\colw}
  \centering
    \includegraphics[width=\linewidth]{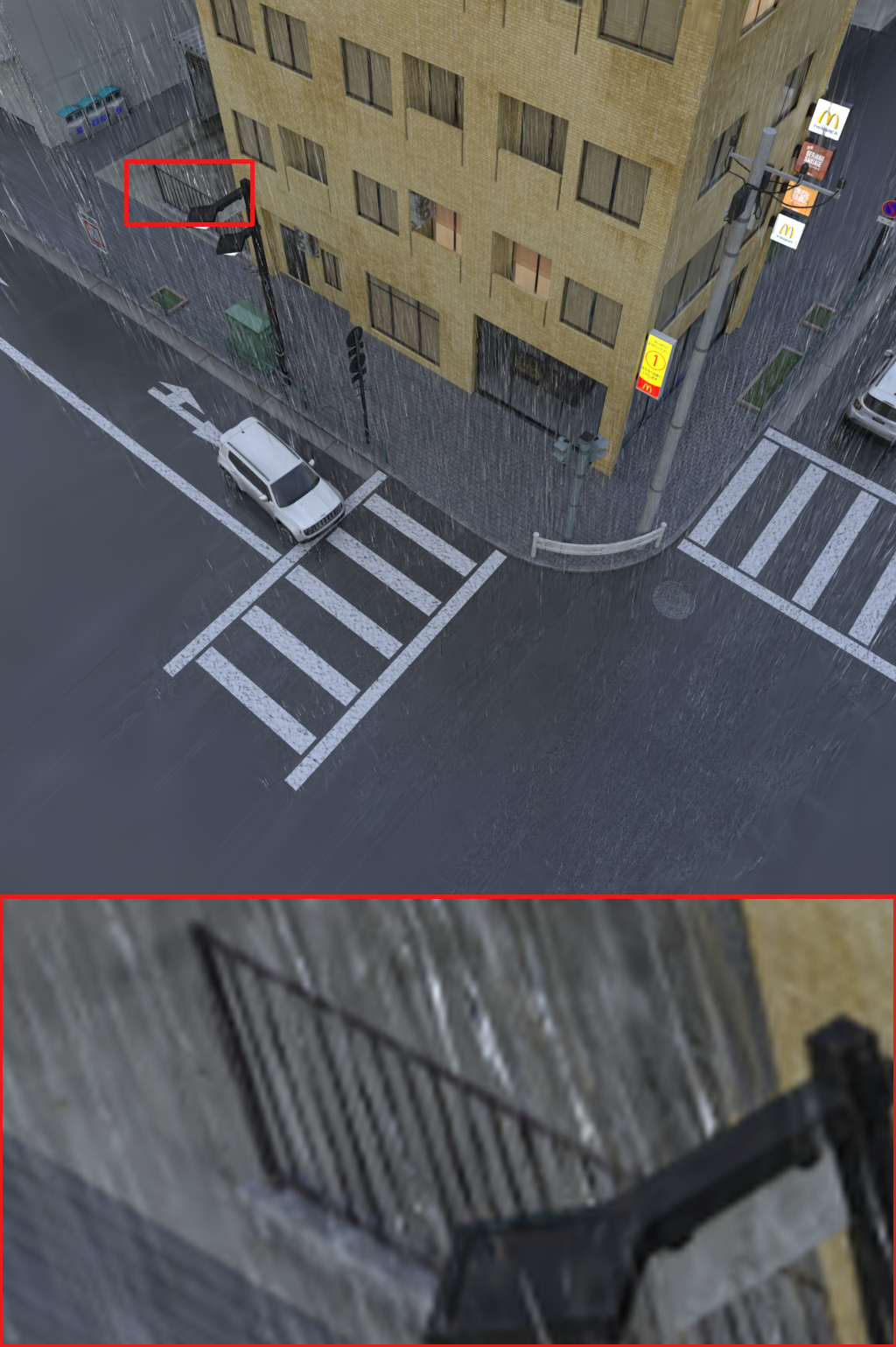}   
    {\small RobustSplat~\cite{2025RobustSplat}}
  \end{minipage}\hspace{\colgap}
  \begin{minipage}[t]{\colw}
  \centering
    \includegraphics[width=\linewidth]{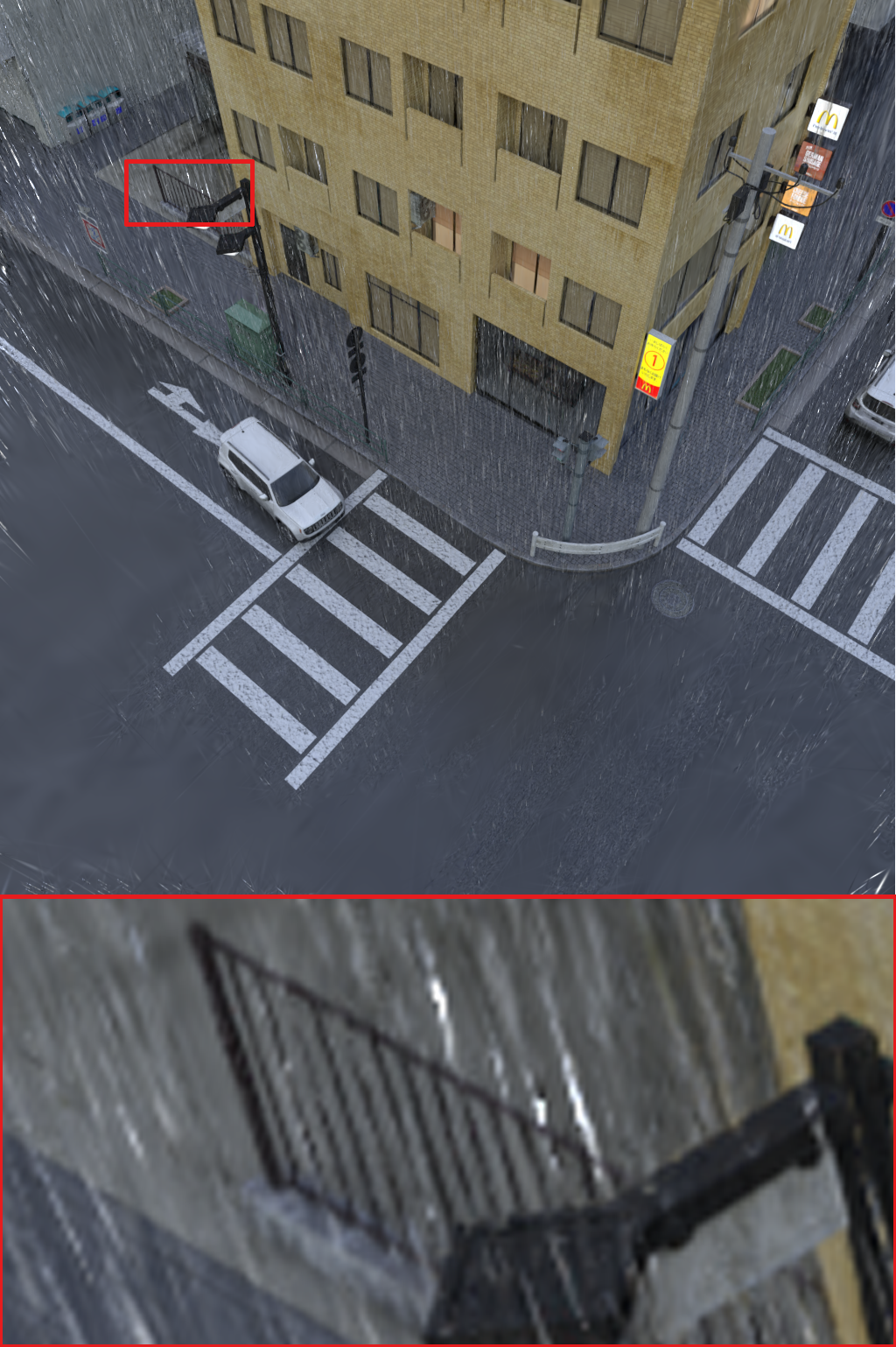}   
    {\small WeatherGS~\cite{qian2025weathergs}}
  \end{minipage}\hspace{\colgap}
  \begin{minipage}[t]{\colw}
  \centering
    \includegraphics[width=\linewidth]{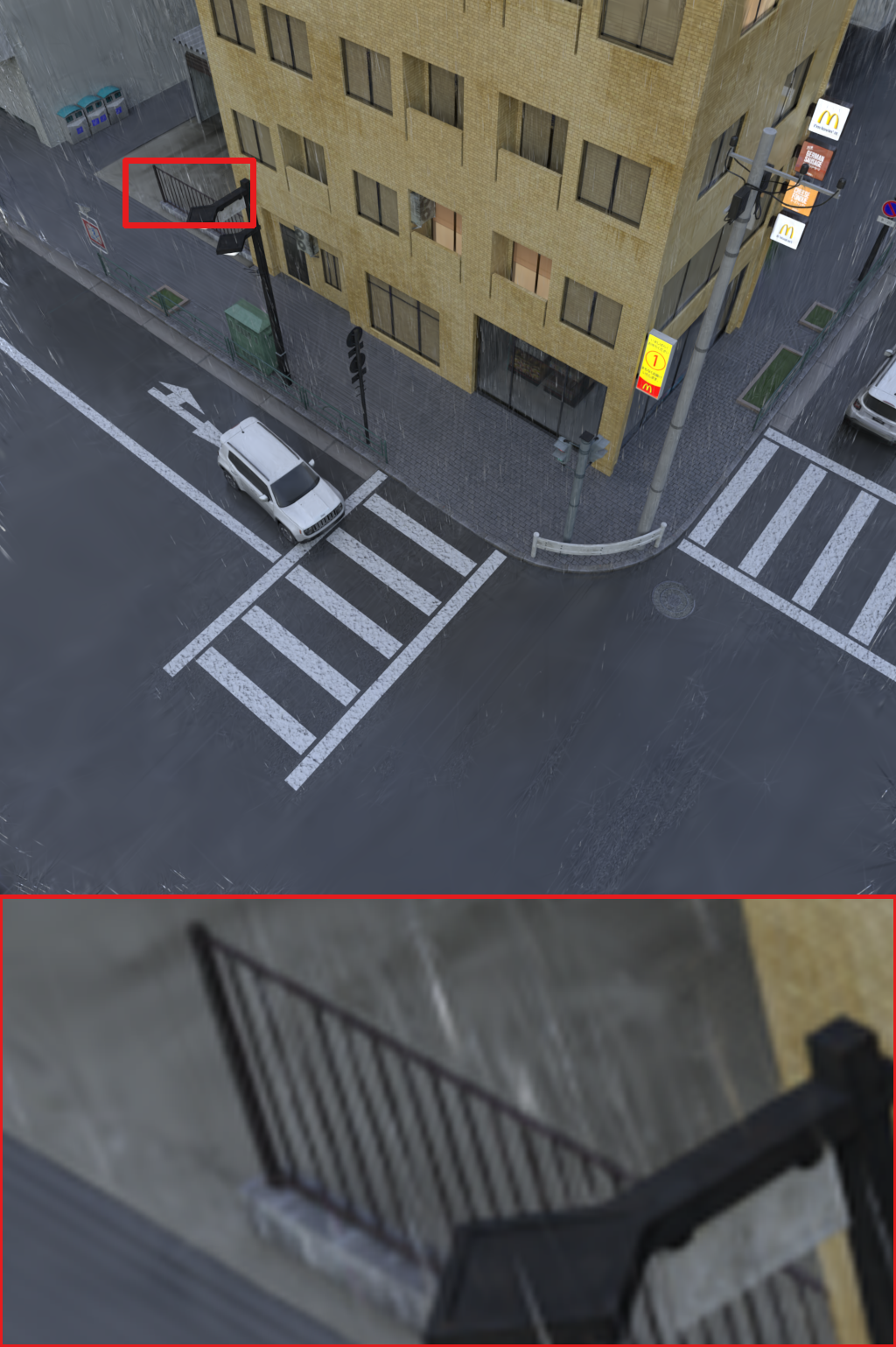}
    {\small RainyScape~\cite{lyu2024rainyscape}}
  \end{minipage}\hspace{\colgap}
  \begin{minipage}[t]{\colw}
  \centering
    \includegraphics[width=\linewidth]{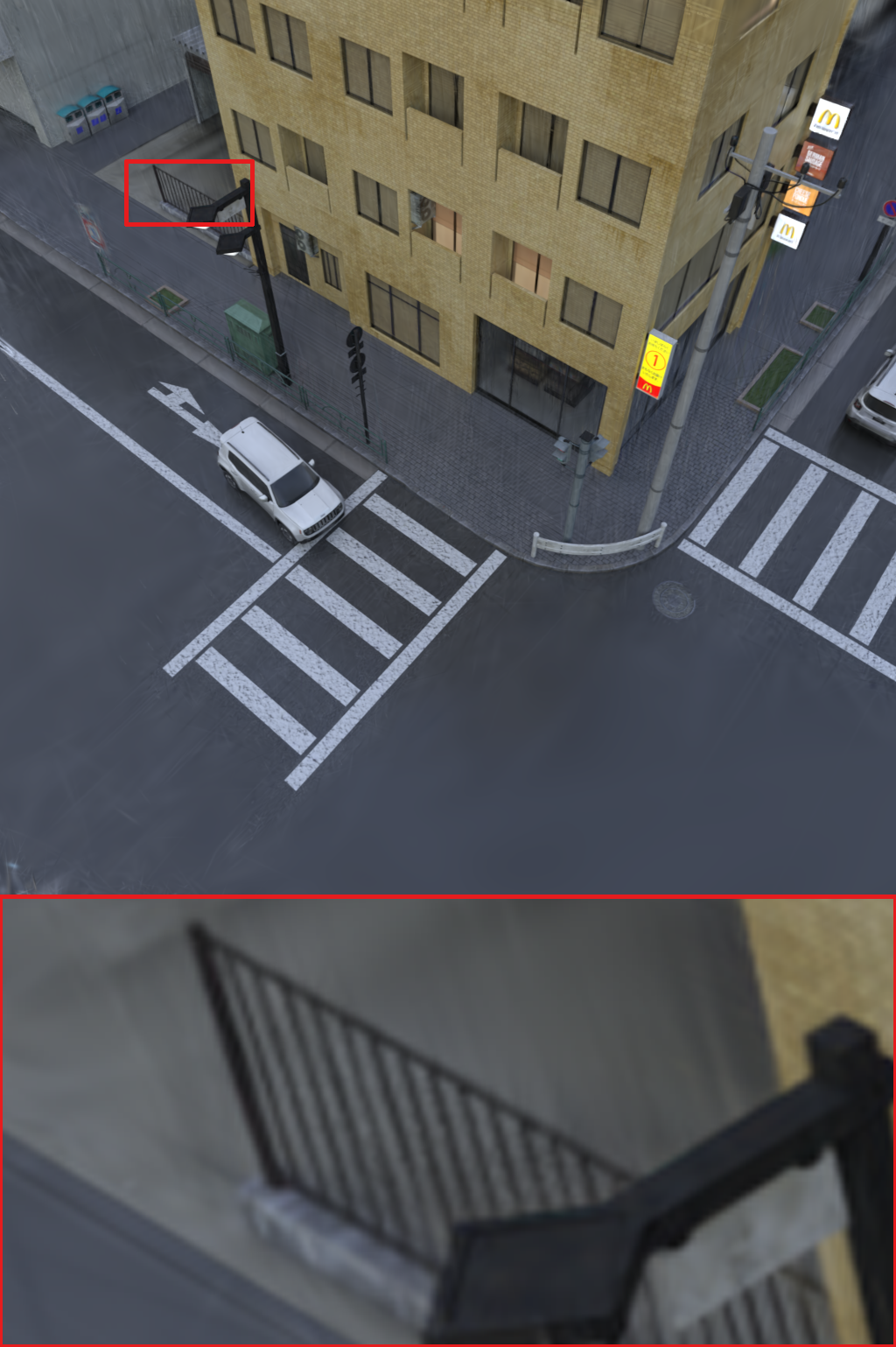}
    {\small Ours}
  \end{minipage}\hspace{\colgap}
  \begin{minipage}[t]{\colw}
  \centering
    \includegraphics[width=\linewidth]{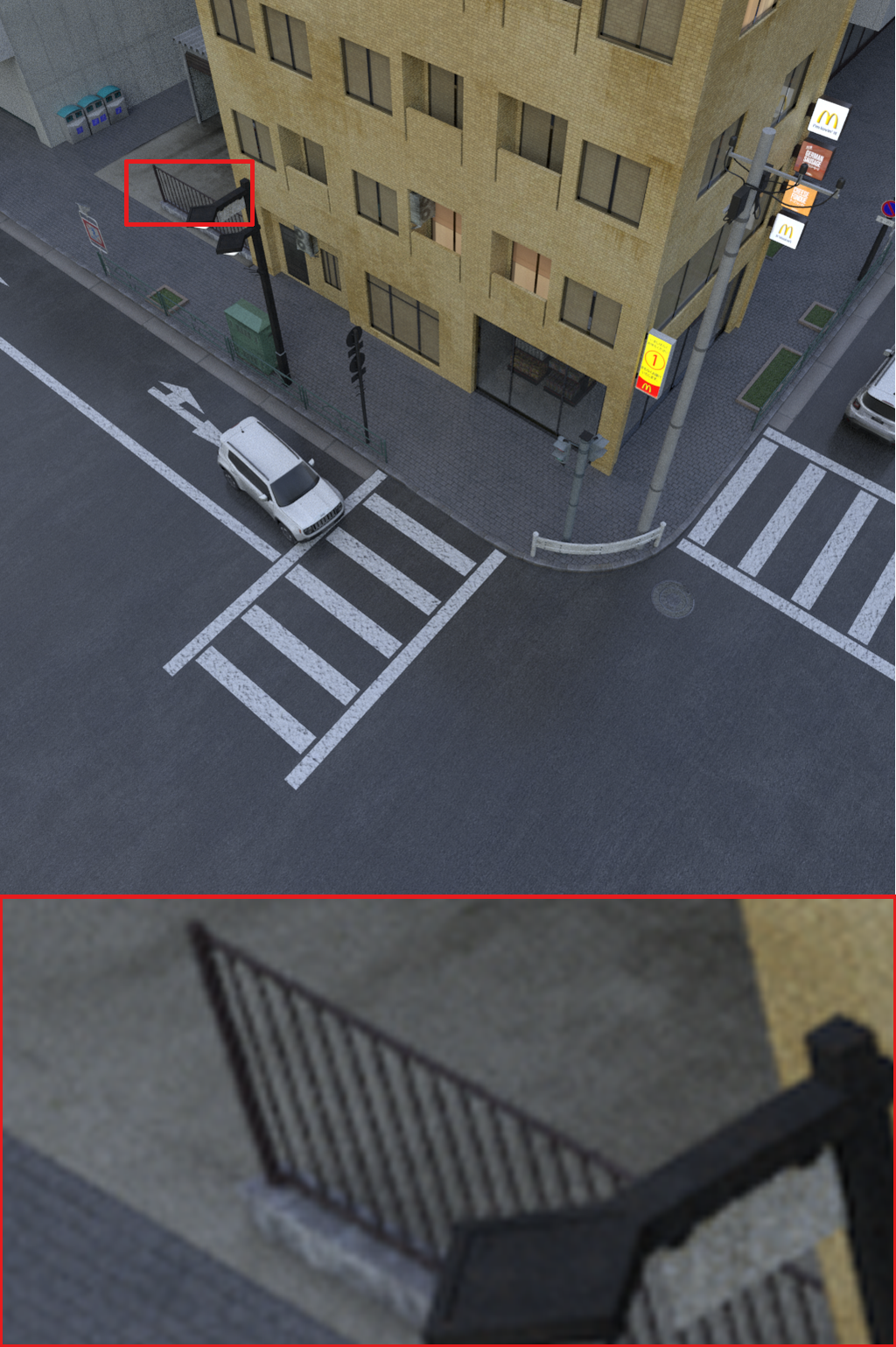}   
    {\small Ground Truth}
  \end{minipage}
  \vspace{-2mm}\caption{Qualitative results on rainy scenes. Best viewed zoomed in.}\vspace{-4mm}
  \label{fig:rain}  
\end{figure*}
\endgroup

\begin{table*}[!t]
\centering
\setlength{\tabcolsep}{3.6pt}
\resizebox{\linewidth}{!}{
\begin{tabular}{l ccc ccc ccc ccc ccc}
\toprule
\multirow{2}{*}{Method} &
\multicolumn{3}{c}{Bicycle} &
\multicolumn{3}{c}{Garden} &
\multicolumn{3}{c}{Factory} &
\multicolumn{3}{c}{Tanabata} &
\multicolumn{3}{c}{Average}\\
\cmidrule(lr){2-4}\cmidrule(lr){5-7}\cmidrule(lr){8-10}\cmidrule(lr){11-13}\cmidrule(lr){14-16}
& PSNR$\uparrow$ & SSIM$\uparrow$& LPIPS$\downarrow$
& PSNR$\uparrow$ & SSIM$\uparrow$& LPIPS$\downarrow$
& PSNR$\uparrow$ & SSIM$\uparrow$& LPIPS$\downarrow$
& PSNR$\uparrow$ & SSIM$\uparrow$& LPIPS$\downarrow$
& PSNR$\uparrow$ & SSIM$\uparrow$& LPIPS$\downarrow$ \\
\midrule
DerainNeRF~\cite{li2024derainnerf}
& 16.08 & 0.581 & 0.632
& 18.83 & 0.435 & 0.421
& 19.96 & 0.669 & 0.194
& 20.07 & 0.632 & 0.379
& 18.73 & 0.579 & 0.406 \\
RobustSplat~\cite{2025RobustSplat}
& 19.17 & 0.622 & 0.326
& 22.40 & 0.693 & 0.270
& 20.76 & 0.797 & 0.200
& 22.86 & \second{0.795} & 0.193
& 21.29 & 0.726 & 0.247 \\
WeatherGS~\cite{qian2025weathergs}
& 17.50 & 0.547 & 0.383
& 21.04 & 0.604 & 0.270
& 19.89 & 0.601 & 0.331
& 21.16 & 0.677 & 0.317
& 19.89 & 0.607 & 0.325 \\
RainyScape~\cite{lyu2024rainyscape}
& \second{19.33} & \second{0.632} & \second{0.270}
& \second{23.83} & \second{0.714} & \second{0.212}
& \second{22.64} & \second{0.832} & \second{0.137}
& \second{23.57} & 0.792 & \second{0.156}
& \second{22.34} & \second{0.742} & \second{0.193} \\
\textbf{Ours}
& \first{20.31} & \first{0.671} & \first{0.249}
& \first{24.21} & \first{0.739} & \first{0.187}
& \first{24.06} & \first{0.867} & \first{0.103}
& \first{24.52} & \first{0.841} & \first{0.112}
& \first{23.27} & \first{0.779} & \first{0.162} \\
\bottomrule
\end{tabular}}\vspace{-2mm}
\caption{Quantitative comparisons on snowy scenes. The \colorbox{red!25}{best} and \colorbox{orange!25}{second-best} scores are color-encoded for clarity.}\vspace{-2mm}\label{tab:desnow}
\end{table*}

\newcommand{\colw}{0.19\linewidth} 
\setlength{\tabcolsep}{2pt}
\begin{figure*}[!t]
  \centering
  \begin{minipage}[t]{\colw}
  \centering
    \includegraphics[width=\linewidth]{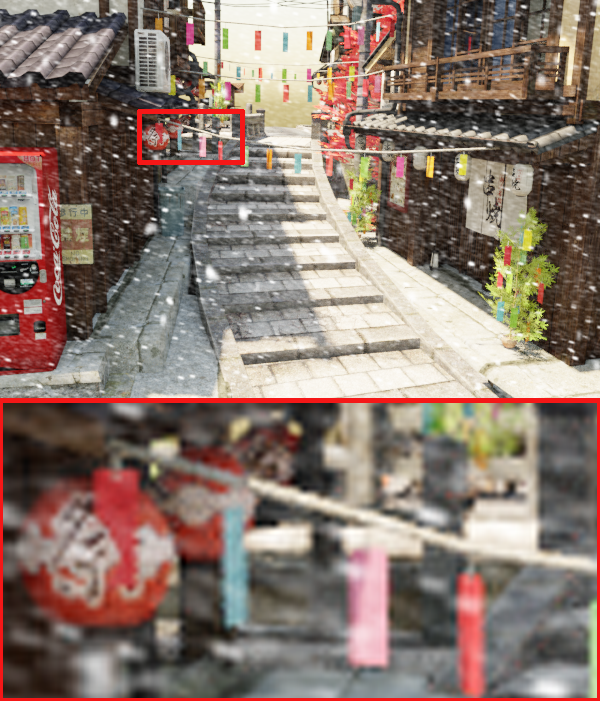}
    {\small Input}
  \end{minipage}\hspace{1pt}
  \begin{minipage}[t]{\colw}
  \centering    
    \includegraphics[width=\linewidth]{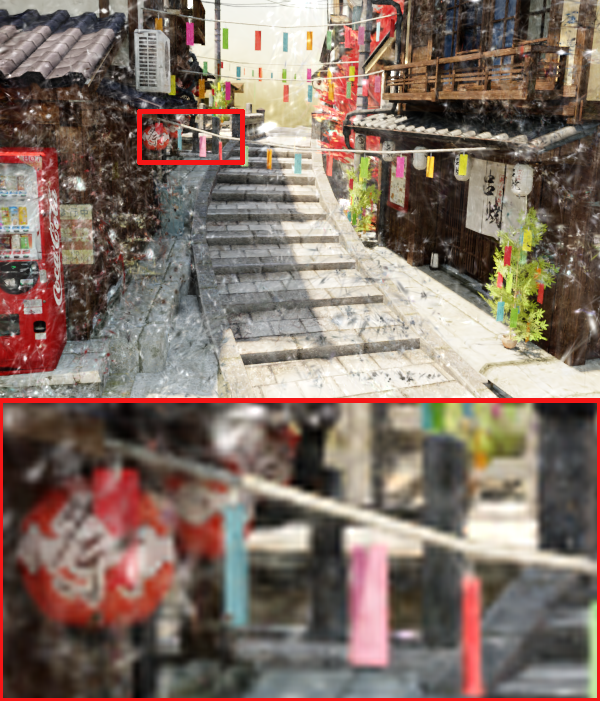}
    {\small RobustSplat~\cite{2025RobustSplat}}
  \end{minipage}\hspace{1pt}
  \begin{minipage}[t]{\colw}
  \centering
    \includegraphics[width=\linewidth]{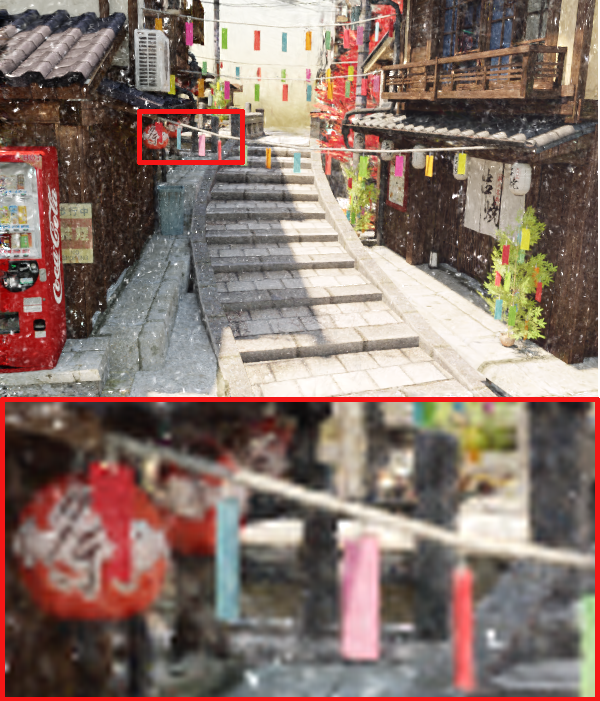}
    {\small WeatherGS~\cite{qian2025weathergs}}
  \end{minipage}\hspace{1pt}
  \begin{minipage}[t]{\colw}
  \centering
    \includegraphics[width=\linewidth]{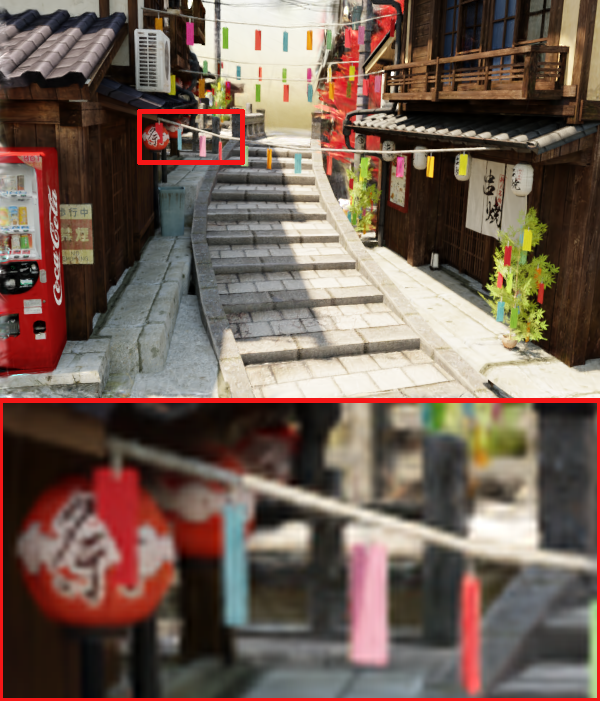}
    {\small Ours}
  \end{minipage}\hspace{1pt} 
  \begin{minipage}[t]{\colw}\centering
    \includegraphics[width=\linewidth]{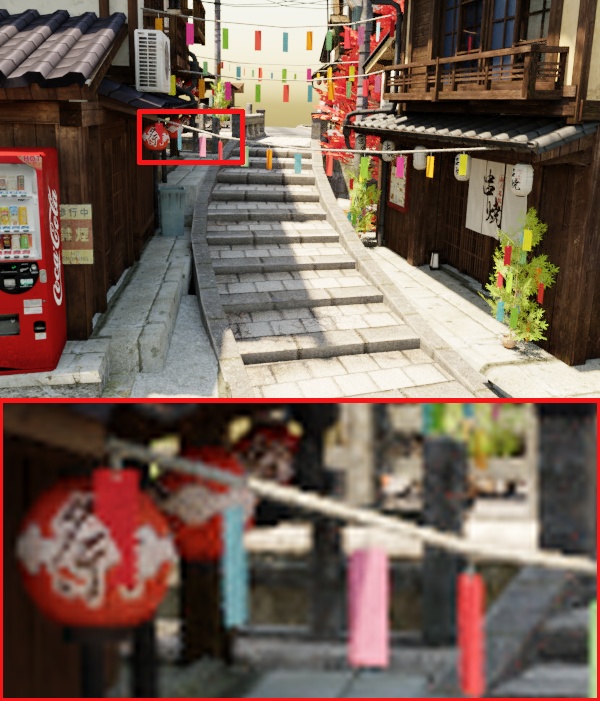}
    {\small Ground Truth}
  \end{minipage}\vspace{-2mm} 
  \caption{Qualitative results on snowy scenes. Best viewed zoomed in.}\vspace{-2mm}
  \label{fig:snow} 
\end{figure*}

\begin{table*}[!t]
\centering
\setlength{\tabcolsep}{3.6pt}
\resizebox{\linewidth}{!}{
\begin{tabular}{l ccc ccc ccc ccc ccc}
\toprule
\multirow{2}{*}{Method} &
\multicolumn{3}{c}{Bicycle} &
\multicolumn{3}{c}{Garden} &
\multicolumn{3}{c}{Factory} &
\multicolumn{3}{c}{Tanabata} &
\multicolumn{3}{c}{Average}\\
\cmidrule(lr){2-4}\cmidrule(lr){5-7}\cmidrule(lr){8-10}\cmidrule(lr){11-13}\cmidrule(lr){14-16}
& PSNR$\uparrow$ & SSIM$\uparrow$& LPIPS$\downarrow$
& PSNR$\uparrow$ & SSIM$\uparrow$& LPIPS$\downarrow$
& PSNR$\uparrow$ & SSIM$\uparrow$& LPIPS$\downarrow$
& PSNR$\uparrow$ & SSIM$\uparrow$& LPIPS$\downarrow$
& PSNR$\uparrow$ & SSIM$\uparrow$& LPIPS$\downarrow$ \\
\midrule
3DGS~\cite{kerbl3Dgaussians}
& 15.71 & 0.473 & 0.370
& 15.86 & 0.588 & 0.327
& 15.28 & 0.681 & 0.240
& 14.30 & 0.500 & 0.392
& 15.29 & 0.561 & 0.332 \\
RobustSplat~\cite{2025RobustSplat}
& 15.93 & 0.474 & 0.375
& 15.98 & 0.579 & 0.338
& 15.40 & 0.688 & 0.230
& 14.98 & 0.531 & 0.365
& 15.57 & 0.568 & 0.327 \\
WeatherGS~\cite{qian2025weathergs}
& 15.96 & 0.489 & 0.367
& 15.65 & 0.574 & 0.469
& 15.75 & 0.509 & 0.416
& 14.88 & 0.531 & 0.430
& 15.56 & 0.526 & 0.420 \\
OR~\cite{guo2024onerestore}+3DGS~\cite{kerbl3Dgaussians}
& \second{17.12} & \second{0.529} & \second{0.366}
& \second{20.18} & \second{0.618} & \second{0.324}
& \second{23.53} & \second{0.836} & \second{0.177}
& \second{19.51} & \second{0.659} & \second{0.299}
& \second{20.09} & \second{0.661} & \second{0.292} \\
\textbf{Ours}
& \first{19.74} & \first{0.603} & \first{0.327}
& \first{22.12} & \first{0.708} & \first{0.209}
& \first{24.02} & \first{0.880} & \first{0.099}
& \first{23.12} & \first{0.777} & \first{0.173}
& \first{22.25} & \first{0.742} & \first{0.202} \\
\bottomrule
\end{tabular}}\vspace{-2mm}
\caption{Quantitative comparisons on hybrid-weather scenes (haze, rain, and snow). The \colorbox{red!25}{best} and \colorbox{orange!25}{second-best} scores are color-encoded.}\vspace{-3mm}\label{tab:mix}
\end{table*}

\begin{figure*}[t]
\centering
\setlength{\tabcolsep}{1pt}
\renewcommand{\arraystretch}{0}
\begin{tabular}{cccccc}
\multicolumn{6}{c}{} \\[2pt]
\includegraphics[width=0.16\linewidth]{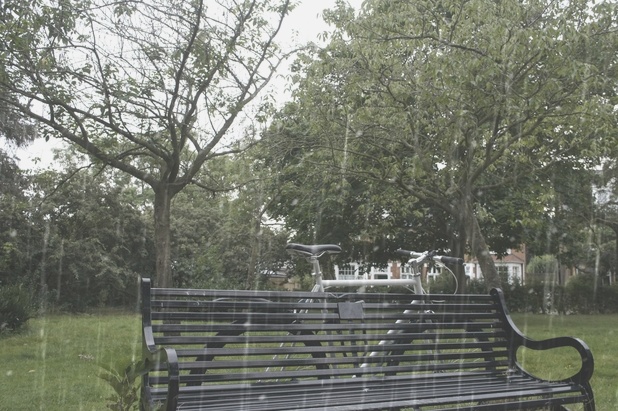} &
\includegraphics[width=0.16\linewidth]{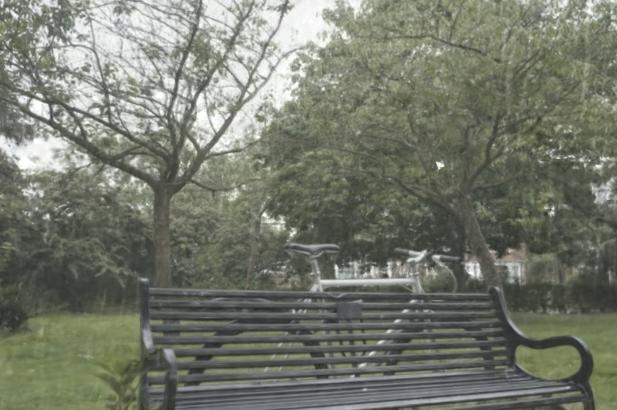} &
\includegraphics[width=0.16\linewidth]{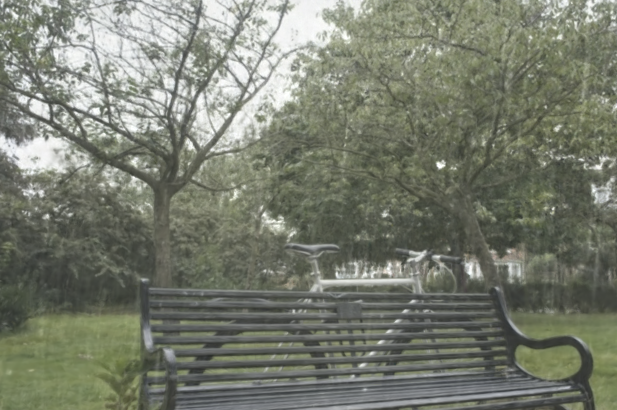} &
\includegraphics[width=0.16\linewidth]{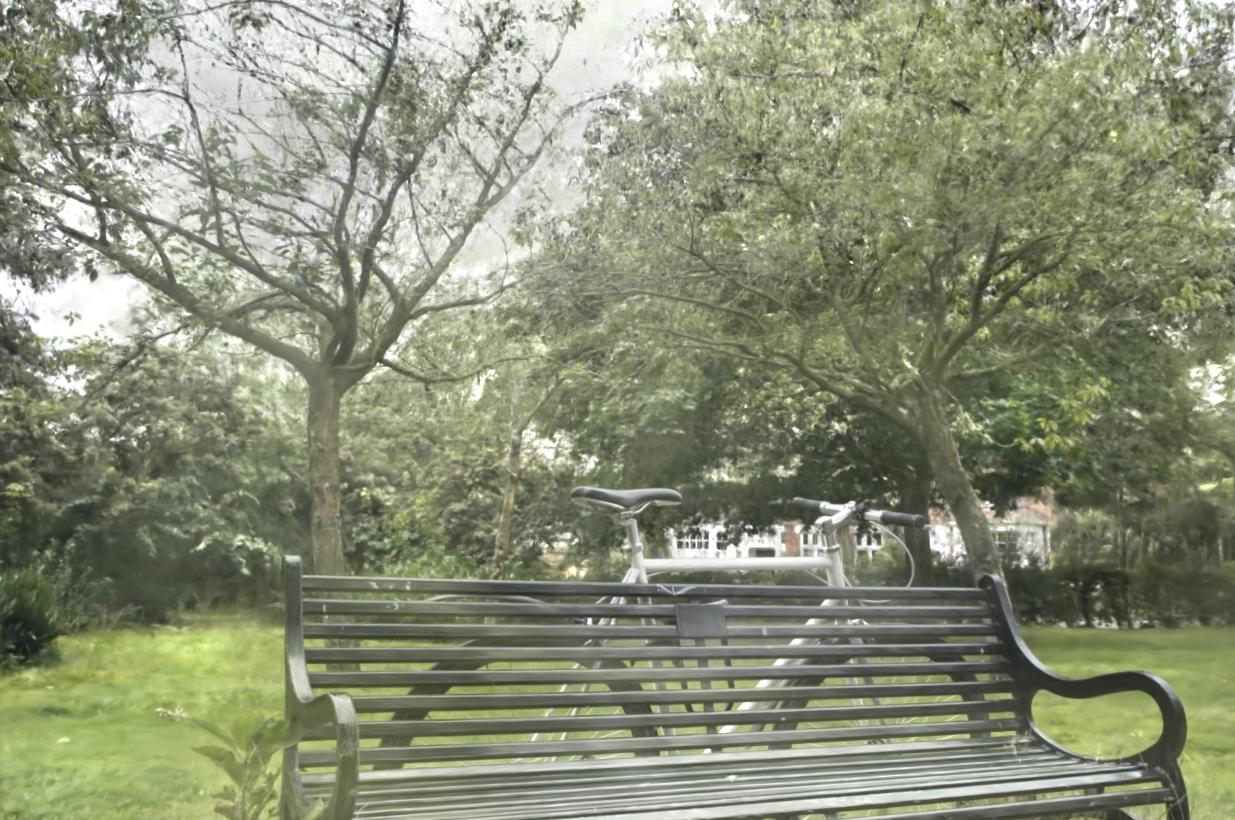} &
\includegraphics[width=0.16\linewidth]{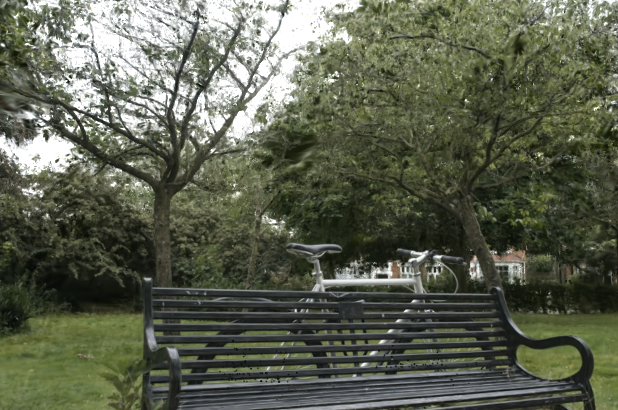} &
\includegraphics[width=0.16\linewidth]{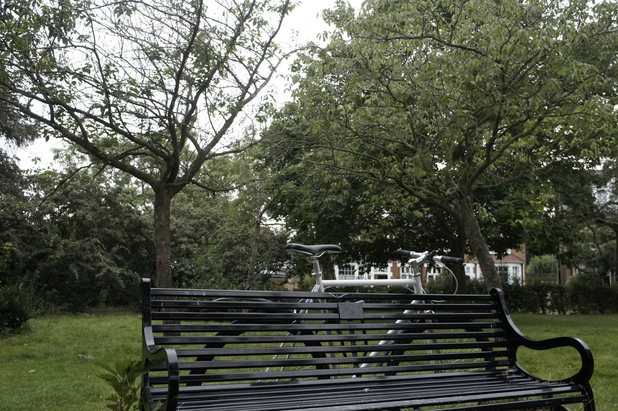} \\[0.5pt]
\multicolumn{6}{c}{} \\[2pt]
\includegraphics[width=0.16\linewidth]{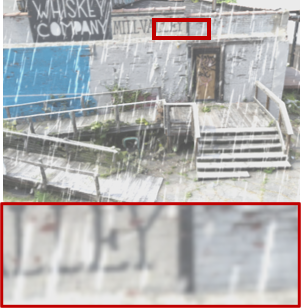} &
\includegraphics[width=0.16\linewidth]{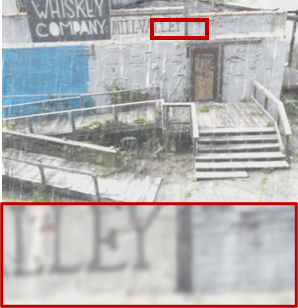} &
\includegraphics[width=0.16\linewidth]{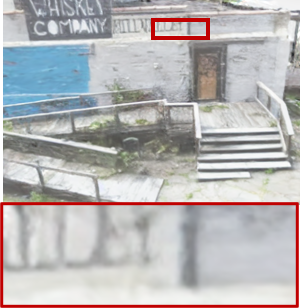} &
\includegraphics[width=0.16\linewidth]{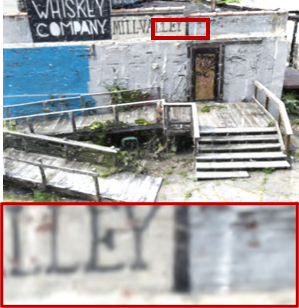} &
\includegraphics[width=0.16\linewidth]{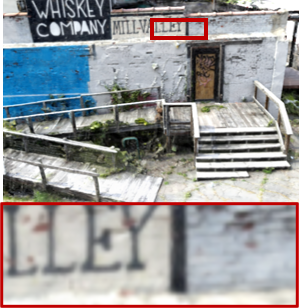} &
\includegraphics[width=0.16\linewidth]{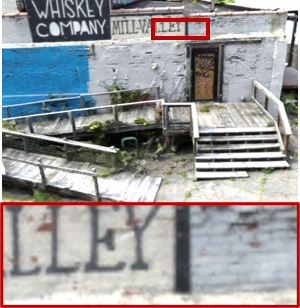} \\[4pt]
\small Input & \small RobustSplat~\cite{2025RobustSplat} & \small WeatherGS~\cite{qian2025weathergs} & \small \parbox[t]{0.16\linewidth}{\centering OR~\cite{guo2024onerestore}+3DGS~\cite{kerbl3Dgaussians}} & \small Ours & \small Ground Truth
\end{tabular}\vspace{-2mm}
\caption{Qualitative results on hybrid-weather scenes (haze, rain, and snow). Best viewed zoomed in.}\vspace{-4mm}
\label{fig:mixed_qual}
\end{figure*}

\subsection{Experimental Setup}
\noindent\textbf{Implementation Details.}
All experiments are conducted on a single NVIDIA RTX 3090 GPU. We adopt a geometry initialization stage for the first 4k iterations, followed by 26k iterations of joint optimization for all learnable components, during which the particulate layer is refreshed every 100 iterations. The sampling count $K$ is set to 64. The reference radius $r_0$ is fixed to 3 pixels, and the loss weight $\lambda_{\mathrm{r}}$ is set to 0.4.

\noindent\textbf{Datasets.}
We evaluate NimbusGS on four adverse-weather tasks: hazy, rainy, snowy, and hybrid scenes. For the hazy, snowy, and hybrid settings, we use four outdoor scenes from Deblur-NeRF~\cite{ma2022deblur} and Mip-NeRF360~\cite{barron2022mip}, including Factory, Tanabata, Garden, and Bicycle, and synthesize weather degradations using the OneRestore~\cite{guo2024onerestore} model. For the rainy task, we use the synthetic rain scenes in RainyScape~\cite{lyu2024rainyscape}. Note that hybrid weather is formed by jointly combining haze, rain, and snow effects.

\noindent\textbf{Baselines and Metrics.}
We compare NimbusGS with radiance-field reconstruction methods that provide official code and can operate under weather-degraded conditions. For hazy scenes, we include SeaSplat~\cite{yang2025seasplat}, WaterSplatting~\cite{li2025watersplatting}, and 3DGS~\cite{kerbl3Dgaussians}. For rainy and snowy scenes, we compare against DerainNeRF~\cite{li2024derainnerf}, RainyScape~\cite{lyu2024rainyscape}, and WeatherGS~\cite{qian2025weathergs}, and also include RobustSplat~\cite{2025RobustSplat} for its ability to model transient effects. For hybrid weather, we use 3DGS, RobustSplat, and WeatherGS, and also consider a restoration-reconstruction baseline where hybrid-degraded inputs are first processed by OneRestore~\cite{guo2024onerestore} and then reconstructed with 3DGS. Performance is evaluated using PSNR, SSIM, and LPIPS against clean reference images. All competitor results are produced with their official implementations with default configurations.

\subsection{Comparison with State-of-the-Art Methods}
\noindent\textbf{Hazy Scenes.}
Quantitative results are reported in Tab.~\ref{tab:dehaze}. NimbusGS achieves the best performance across all metrics, surpassing the second-best competitor by 4.64 dB in PSNR and 0.067 in SSIM on average. Visual comparisons in Fig.~\ref{fig:fog} show that existing methods often struggle to maintain consistent brightness and faithful colors, leading to uneven darkening and distortions. In contrast, NimbusGS preserves both luminance and fine textures more effectively, benefiting from the improved decoupling between continuous scattering and appearance.

\noindent\textbf{Rainy Scenes.}
Tab.~\ref{tab:derain} summarizes the performance of NimbusGS and the comparison methods. NimbusGS clearly outperforms the rainy-specific baselines, achieving average gains of 2.54 dB in PSNR and 0.053 in SSIM. The qualitative results in Fig.~\ref{fig:rain} show that our method yields cleaner renderings with sharper boundaries.

\noindent\textbf{Snowy Scenes.}
Tab.~\ref{tab:desnow} shows that NimbusGS significantly outperforms all comparison methods across each evaluation metric. This advantage comes from our particulate layer modeling, which absorbs bright snow particles and suppresses floating artifacts in novel views. The qualitative comparisons in Fig.~\ref{fig:snow} further show that NimbusGS reconstructs snow-degraded scenes with clearer textures and more faithful structures.

\noindent\textbf{Hybrid Weather Scenes.} 
Tab.~\ref{tab:mix} reports that \mbox{NimbusGS} delivers the best overall performance under hybrid weather. The restoration–reconstruction pipeline represented by OneRestore+3DGS (denoted as OR+3DGS) often exhibits geometric and appearance drift across views, while task-specific 3D methods such as 3DGS-based approaches, RobustSplat, and WeatherGS struggle to jointly handle medium attenuation and transient particles. By modeling continuous scattering and particulate effects within a unified differentiable framework, NimbusGS produces cleaner distant geometry and fewer view-dependent artifacts, as shown in Fig.~\ref{fig:mixed_qual}. Additional visual comparisons for all scenes are provided in the supplementary material.

\subsection{Ablation Study}
\noindent\textbf{Effectiveness of CSM.} 
We evaluate CSM using hazy-scene data, with results reported in the top block of Tab.~\ref{tab:ablation}. The first variant, Gaussian-Driven Extinction, removes the volumetric extinction field and instead assigns the extinction coefficient directly to each Gaussian primitive. The degraded performance of this variant shows that, when extinction, geometry, and appearance are parameterized within the same Gaussian field and optimized together, their attributes become tightly coupled. This coupling hampers the separation of medium effects from scene structure, causing unstable decomposition and reducing reconstruction quality. 

The second variant, Uniform Extinction, replaces the spatially varying extinction with a global constant. Although simpler, this setting cannot capture the spatial and directional variations of real haze and skylight, often leading to spatially inconsistent dehazing and noticeable reconstruction bias. In contrast, CSM achieves the best reconstruction quality. Its decoupled and spatially adaptive formulation enables stable optimization and maintains strong cross-view consistency.

\begin{table}[t]
\centering
\setlength{\tabcolsep}{3.6pt}           
\renewcommand{\arraystretch}{1.08}    
\resizebox{\linewidth}{!}{
\begin{tabular}{ccccc}   
\toprule
Variant & \multicolumn{1}{|c|}{Type} & PSNR$\uparrow$ & SSIM$\uparrow$ & LPIPS$\downarrow$ \\
\midrule
Gaussian-Driven Extinction  & \multicolumn{1}{|c|}{\multirow{3}{*}{Hazy}} & 19.29 & 0.709 & 0.284 \\
Uniform Extinction  & \multicolumn{1}{|c|}{} & 20.55 & 0.749 &  0.211 \\
\textbf{Ours}  &  \multicolumn{1}{|c|}{} & \textbf{20.79} & \textbf{0.776} & \textbf{0.190}  \\
\cmidrule(lr){1-5}
Restored-Derived Residuals   & \multicolumn{1}{|c|}{\multirow{2}{*}{Rainy}} & 29.65 & 0.844 & 0.227 \\
\textbf{Ours} & \multicolumn{1}{|c|}{} & \textbf{32.41} & \textbf{0.904} & \textbf{0.139} \\
\cmidrule(lr){1-5}
w/o GGS & \multicolumn{1}{|c|}{\multirow{2}{*}{Hybrid}}        & 21.86 & 0.721 & 0.237 \\
\textbf{Ours} & \multicolumn{1}{|c|}{} & \textbf{22.25} & \textbf{0.742} & \textbf{0.202} \\
\bottomrule
\end{tabular}}
\vspace{-2mm}\caption{Ablation study of CSM (top block), PLM (middle block), and GGS (bottom block) under hazy, rainy, and hybrid-weather settings. The best results are marked in \textbf{bold}.}\vspace{-4mm}
\label{tab:ablation}
\end{table}

\noindent\textbf{Effectiveness of PLM.} 
We evaluate PLM on rainy-scene data. In this variant, the cross-view–consistent static image $I_\text{con}$ used for residual extraction is replaced by pseudo-clean images generated by the all-in-one restoration model OneRestore. However, as a 2D model, OneRestore performs single-view correction without enforcing cross-view consistency. The resulting inconsistencies propagate into the residuals and introduce view-dependent errors that conflict with the static geometry, leading to floating artifacts, geometric drift, and degraded reconstruction quality. The inferior performance of this variant underscores the importance of keeping transient effects in a view-dependent pixel space and decoupled from static geometry.

\noindent\textbf{Effectiveness of GGS.} 
We evaluate the role of GGS on hybrid-weather scenes by removing it and reverting to the default 3DGS setting, where densification is triggered directly by the raw, unscaled gradient. Under weather degradation, distant surfaces produce extremely weak gradients, making this default trigger unreliable and leaving the Gaussian field sparse in far-range regions. As a result, distant geometry becomes under-densified, yielding blurred structures and missing fine details. With GGS, these weak-gradient regions receive amplified and more reliable densification cues, allowing distant structures to accumulate sufficient Gaussian support during optimization. This leads to sharper and more complete far-range geometry, more stable shape and texture, and overall better reconstruction quality. Additional visual comparisons and extended ablation analyses are provided in the supplementary material.

\section{Conclusion}
We propose NimbusGS, a unified framework for weather-robust 3D scene reconstruction. By decomposing weather effects into a continuous transmission field and per-view particle residuals, and stabilizing optimization under visibility imbalance, NimbusGS achieves clean and reliable geometry across diverse conditions. Experiments show that NimbusGS not only significantly outperforms task-specific baselines but also attains state-of-the-art performance in hybrid-weather scenarios, establishing a new direction for 3D reconstruction in adverse real-world environments.

\clearpage
\setcounter{page}{1}
\maketitlesupplementary

\section{Additional Implementation Details}
\subsection{Training Configuration}
During the geometry initialization stage, the learning rate of the Gaussian scale parameters is reduced to $2\times10^{-3}$ for stable early optimization, while the other learning rate settings follow 3DGS. In the joint optimization stage, all Gaussian attributes revert to the default 3DGS settings.

For the extinction field $\beta$, we build an axis-aligned bounding box (AABB) from the Gaussian centers and enlarge it by $2\times$ along all axes so that the entire scene and all valid ray segments fall within the volume. A uniform voxel grid of size $128^3$ is constructed inside this AABB, with each voxel initialized as a single scalar drawn from a Gaussian distribution (mean = 0, std = 0.01). The extinction field and the scattering-color MLP are optimized with learning rates of $5\times10^{-3}$ and $5\times10^{-4}$, respectively.

Details regarding ray sampling and volumetric rendering are provided in the main paper and omitted here for brevity.
\subsection{Adverse-weather Degradation Synthesis}
Since no public dataset provides multi-view scenes with controllable hybrid adverse-weather effects, we generate the required degradations using the composite model of OneRestore~\cite{guo2024onerestore}. Clean images are taken from the outdoor scenes of Mip-NeRF360~\cite{barron2022mip} and Deblur-NeRF~\cite{ma2022deblur}. Haze (H) is synthesized using the standard atmospheric scattering model, where the extinction coefficient and airlight intensity are randomly sampled within the ranges defined by the OneRestore implementation. Depth maps are predicted by the pretrained Depth-Anything v2~\cite{yang2024depth} network and used to compute transmission. Rain and snow components follow the public OneRestore pipeline: rain streaks are sampled from RainStreakGen~\cite{garg2006photorealistic}, and snow masks are taken from Snow100K~\cite{liu2018desnownet} and blended with the clean images. This procedure yields six degradation types: haze (H), snow (S), rain+snow (R+S), rain+haze (R+H), snow+haze (S+H), and rain+snow+haze (R+S+H).

Note that real-world haze and snow datasets do not provide multi-view scenes or camera poses and are therefore unsuitable for our 3D reconstruction setting, so these weather effects are synthesized. In contrast, the rain (R) setting is directly taken from the multi-view synthetic rainy scenes in the Rainyscape dataset rather than generated with OneRestore.

\begin{algorithm}[!t]
\caption{Training of NimbusGS}
\label{alg:nimbusgs}
\KwIn{\begin{tabular}[t]{@{}l@{}}
scene images $\{I_{\text{in}}\}$, initialized 3D Gaussians $\mathcal{G}$,\\ sampling count $K$, reference radius $r_0$, \\
loss weights $\lambda_r$, iteration numbers $M_{\text{init}}$\\
(geometry initialization) and $M_{\text{joint}}$ (joint\\
optimization), residual refresh interval $Z_{\text{ref}}$.
\end{tabular}}
\KwOut{\begin{tabular}[t]{@{}l@{}}
optimized Gaussians $\mathcal{G}$, extinction field $\beta$, \\
and particulate layer $R$.
\end{tabular}}
\BlankLine
\small\tcp{Stage 1: Geometry Initialization}
\For{$i = 1$ \KwTo $M_{\text{init}}$}{
    Render continuous-medium output $I_{\text{con}}$\;
    Update model parameters using $\mathcal{L}_{\text{ini}}(I_{\text{in}}, I_{\text{con}})$\;
}
Compute initial particulate layer: $R = \mathrm{ReLU}(I_{\text{in}} - I_{\text{con}})$;
\BlankLine
\tcp{Stage 2: Joint Optimization}
\For{$j = 1$ \KwTo $M_{\text{joint}}$}{
    Render degraded output $I_{\text{deg}}$ using CSM and PLM\;    
    Update $\mathcal{G}$ with $\mathcal{L}$ and GGS\;
    Update extinction field $\beta$ and the airlight module by $\mathcal{L}$\;

    \If{$j \bmod Z_{\text{ref}} = 0$}{
        Recompute $I_{\text{con}}$ from current $\mathcal{G}$ and medium components\;
        Refresh particulate layer: $R = \mathrm{ReLU}(I_{\text{in}} - I_{\text{con}})$\;
    }
}
\end{algorithm}

\begin{figure*}[h]
	\centering
	\includegraphics[width=0.98\linewidth]{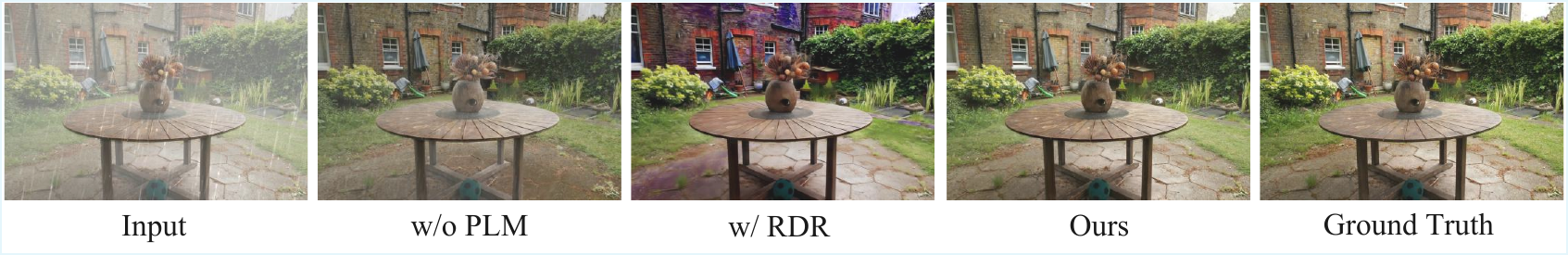}
	\caption{Qualitative ablation of Particulate Layer Modeling. Best viewed zoomed in.}
	\label{fig:plm}
\end{figure*}
\begin{figure}[!t]
	\centering
	\includegraphics[width=\linewidth]{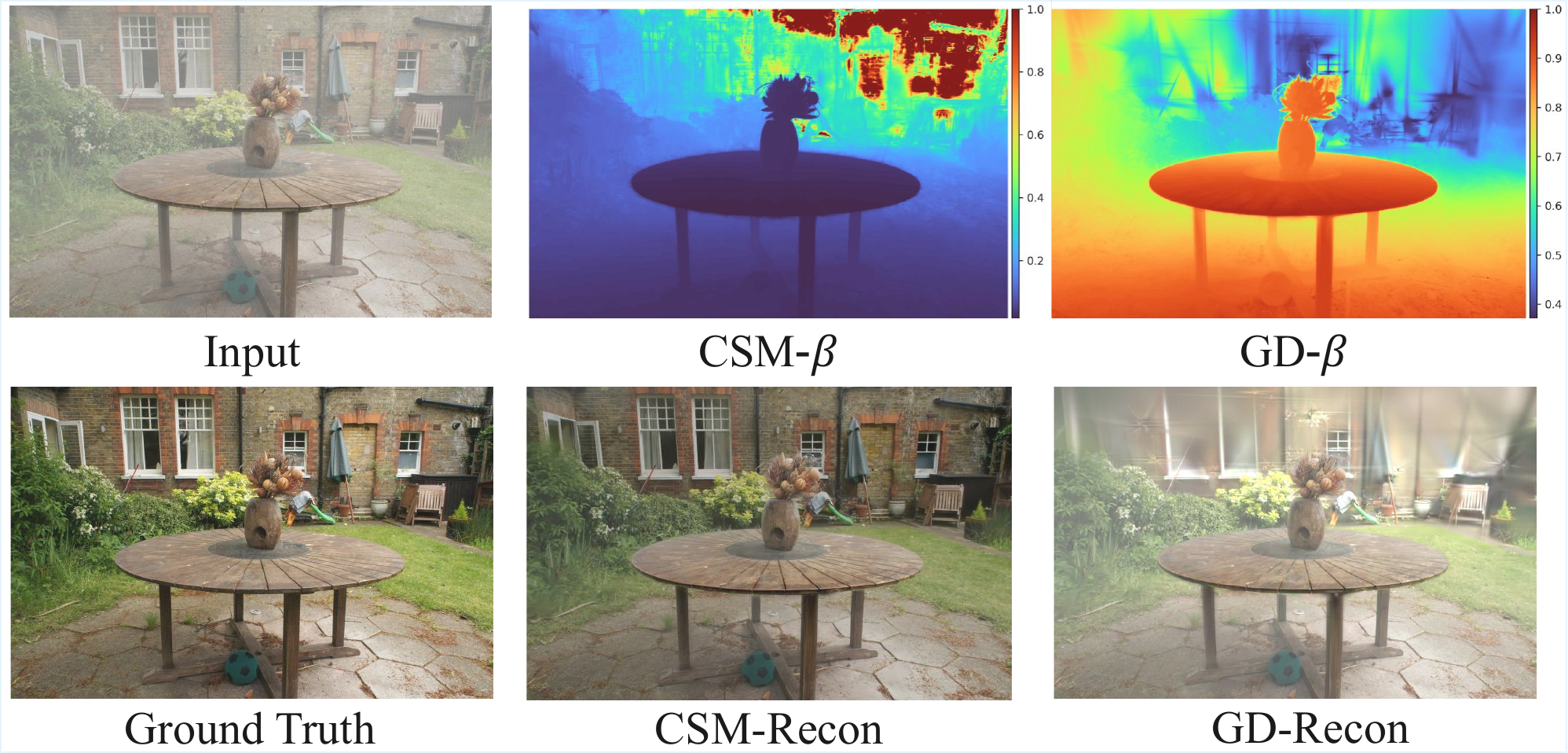}
	\caption{Qualitative ablation of Continuous Scattering Modeling.}
	\label{fig:csm}
\end{figure}
\subsection{Training Pipeline}
We provide the complete training workflow in Algorithm~\ref{alg:nimbusgs}. It offers a compact procedural description for implementing NimbusGS.
\section{Additional Ablation Studies}
\subsection{Continuous Scattering Modeling}

Fig.~\ref{fig:csm} visualizes the ablation of our Continuous Scattering Modeling (CSM) under the haze setting, which offers a clean environment for examining the extinction field. We omit the Uniform Extinction variant since its spatially constant $\beta$ contains no meaningful structure. For the variants that yield spatial variation, we compare the $\beta$ estimated by CSM (CSM-$\beta$) with the Gaussian-Driven Extinction baseline (GD-$\beta$), where $\beta$ is directly optimized as Gaussian parameters, and we also show their corresponding reconstruction results (CSM-Recon and GD-Recon). Each $\beta$ map is normalized independently for visualization because the two approaches follow fundamentally different numerical distributions.

The GD-$\beta$ map shows strong geometry dependence, with high values clustering near bright surfaces and object boundaries and unstable fluctuations in distant regions. This indicates entanglement between $\beta$ and surface radiance, resulting in incomplete haze removal and geometry-coupled artifacts in the GD-Recon output.

In contrast, CSM-$\beta$ exhibits smooth and depth-coherent variation and effectively avoids texture leakage. Foreground regions maintain low extinction, while background areas progressively accumulate higher values consistent with volumetric scattering. The resulting CSM-Recon preserves sharper structures and yields more stable far-field appearance, demonstrating that CSM produces a physically meaningful and depth-coherent extinction field.

\subsection{Particulate Layer Modeling}
Fig.~\ref{fig:plm} provides qualitative examples of PLM. We visualize on the H+R setting, where particle effects are more clearly distinguishable.

Without PLM (w/o PLM), the model is forced to explain both continuous-medium effects and particle streaks using the same volumetric formulation. This entangles the two degradation types and hampers the estimation of the extinction field, leading to incomplete haze removal and leaving noticeable rain streaks in the final reconstruction.

We further evaluate the Restored-Derived Residuals (RDR) alternative (w/ RDR), where an all-in-one restoration method (OneRestore) produces $I_{\text{restore}}$ as an estimate of $I_{\text{con}}$, and the residual is computed as $I_{\text{deg}}-I_{\text{restore}}$. Such all-in-one models tend to suppress multiple degradation types simultaneously, introducing heterogeneous artifacts into the residual. As 2D restorers, they may further produce view-dependent color variations that propagate into the 3D reconstruction and appear as color shifts or overly smoothed regions.

NimbusGS instead renders a multi-view-consistent $I_{\text{con}}$ from the Gaussian field and separates particle effects via PLM, yielding cleaner removal of rain streaks and more stable appearance consistency.
\subsection{Geometry-Guided Gradient Scaling}
\begin{figure*}[!t]
	\centering
	\includegraphics[width=0.9\linewidth]{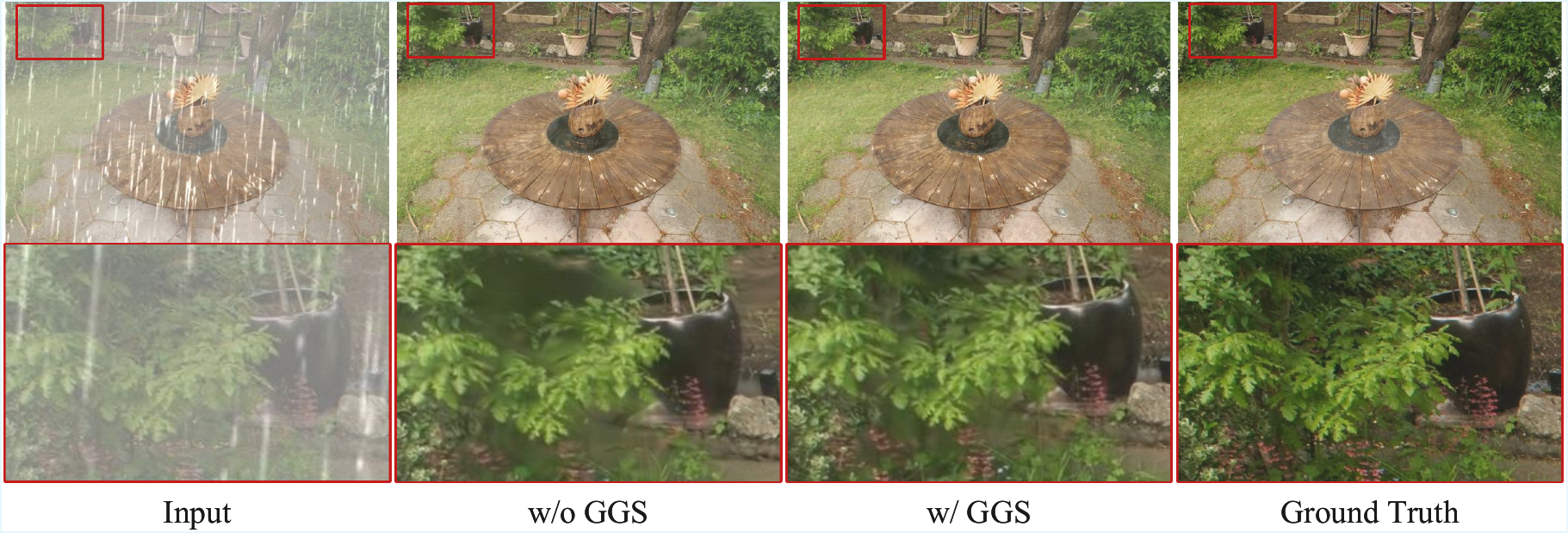}
	\caption{Qualitative ablation of Geometry-Guided Gradient Scaling. Best viewed zoomed in.}
	\label{fig:ggs}
\end{figure*}

We further study the behavior of GGS through qualitative and factor-wise quantitative analyses.

\noindent\textbf{Qualitative Ablation of GGS.}
Fig.~\ref{fig:ggs} shows a comparison on an H+R scene under two settings: w/o GGS and w/ GGS. Without GGS, distant regions suffer from pronounced detail loss and appear overly blurred, mainly due to weakened supervision and the inability of coarse far-field Gaussians to trigger timely densification. With GGS, the depth-, radius-, and error-aware scaling better prioritizes refinement in these challenging areas, yielding sharper geometry and noticeably finer background details.

\noindent\textbf{Factor-wise Quantitative Analysis.}
We further conduct a factor-level ablation on four hybrid-weather scenes (Bicycle, Garden, Factory, and Tanabata) under the H+R+S setting. The averaged results are reported in Tab.~\ref{tab:ablation_ggs}. Removing any single factor from GGS leads to a consistent drop in reconstruction quality, indicating that depth, projected radius, and reconstruction error jointly contribute to effective densification guidance.

Among the three factors, the depth-based component shows the largest impact, as it enhances the supervision of distant structures where weather-induced attenuation makes optimization particularly challenging. The projected-radius component and the reconstruction-error term offer comparable additional gains: the former helps prioritize large Gaussians that dominate far-field regions, while the latter highlights areas that remain under-refined, providing modest but stable improvements across metrics.

\begin{table}[!t]
\centering
\begin{tabular}{lccc}
\toprule
Variant & PSNR$\uparrow$& SSIM$\uparrow$ & LPIPS$\downarrow$\\
\midrule
w/o Depth 
& 21.98 & 0.731 & 0.214 \\
w/o Projected Radius 
& 22.14 & 0.733 & 0.209 \\
w/o Reconstruction Error 
& 22.09 & 0.729 & 0.211 \\
\midrule
\textbf{Ours}
& \textbf{22.25} & \textbf{0.742} & \textbf{0.202} \\
\bottomrule
\end{tabular}
\caption{Factor-wise ablation of GGS. Best results are marked in \textbf{bold}.}
\label{tab:ablation_ggs}
\end{table}

\subsection{Loss Terms} 
\begin{table}[!t]
\centering
\begin{tabular}{lccc}
\toprule
Variant & PSNR↑ & SSIM↑ & LPIPS↓\\
\midrule
w/o $\mathcal L_{\text{DCP}}$
& 17.32 & 0.610 & 0.233 \\
w/ $\mathcal L_{\text{DCP}}$ (stage 2)
& 20.92 & 0.718 & 0.220 \\
w/o $\mathcal L_{\text{TV}}$
& 22.17 & 0.735 & 0.211 \\
\midrule
\textbf{Ours} & \textbf{22.25} & \textbf{0.742} & \textbf{0.202}  \\
\bottomrule
\end{tabular}
\caption{Ablation of loss terms. Best results are marked in \textbf{bold}.}
\label{tab:ablation_loss}
\end{table}

To assess the influence of each loss, we individually remove $\mathcal L_{\text{DCP}}$ and $\mathcal L_{\text{TV}}$, extend $\mathcal L_{\text{DCP}}$ to the second stage, and evaluate these variants on four hybrid-weather scenes (Bicycle, Garden, Factory, and Tanabata) rendered under the H+R+S setting. The averaged results are reported in Tab.~\ref{tab:ablation_loss}. Removing $\mathcal L_{\text{DCP}}$ produces the most significant drop in performance. Although the model can still separate clean appearance from weather scattering through its underlying structure, the absence of this prior weakens this decomposition and leads to noticeably lower reconstruction accuracy across all metrics. However, when $\mathcal L_{\text{DCP}}$ is kept in the second stage, the reconstruction quality also degrades, typically producing darker outputs. This behavior is consistent with the bias of the dark channel prior, which tends to favor locally dark pixels and may therefore drive the optimization toward overly dark solutions in later training. Removing $\mathcal L_{\text{TV}}$ results in a moderate decrease, but introduces small spatial irregularities in the extinction field, which degrade reconstruction stability.

\subsection{Geometry Initialization Schedule} 
We ablate the geometry-initialization iteration count $M_\text{init}$ on two scenes from Mip-NeRF360 (Bicycle and Garden) under the H+R+S setting, where particle-based degradations naturally reveal whether the initialization properly separates static geometry from transient weather artifacts. The averaged results are shown in Tab.~\ref{tab:ablation_iteration}. Very short schedules (\eg, 1K--2K) do not allow the Gaussian field to establish reliable geometry, causing part of the static structure to leak into the particulate residuals and preventing a clean separation between geometry and particle artifacts. Conversely, excessively long schedules (\eg, 10K--15K) cause the Gaussians to absorb particle patterns as if they were static content, leading to floating artifacts and reducing the completeness of the residual map produced by the particulate layer modeling. A setting of 4K iterations achieves the best balance, providing well-formed geometry while keeping particle effects out of the Gaussian representation and consistently delivering the highest reconstruction quality.

\begin{table}[!t]
  \centering
  \resizebox{\linewidth}{!}{
    \begin{tabular}{c ccc ccc}
    \toprule
    \multirow{2}{*}{$M_\text{init}$} &
    \multicolumn{3}{c}{Bicycle} &
    \multicolumn{3}{c}{Garden}\\
    \cmidrule(lr){2-4}\cmidrule(lr){5-7}
     & PSNR$\uparrow$ & SSIM$\uparrow$ & LPIPS$\downarrow$ & PSNR$\uparrow$ & SSIM$\uparrow$ & LPIPS$\downarrow$ \\
    \midrule
    1K  & 17.69 & 0.521 & 0.408 & 20.64 & 0.651 & 0.280\\
    2K  & 18.01 & 0.573 & 0.345 & 21.75 & 0.679 & 0.256\\
    4K  & \textbf{19.74} & \textbf{0.603} & \textbf{0.327} & \textbf{22.12} & \textbf{0.708} & \textbf{0.209}\\
    10K & 19.16 & 0.593 & 0.335 & 22.09 & 0.703 & 0.237 \\
    15K & 19.24 & 0.591 & 0.334 & 22.07 & 0.702 & 0.241\\
    \bottomrule
  \end{tabular}}  
  \caption{Ablation of geometry-initialization iteration count. Best results are marked in \textbf{bold}.}
  \label{tab:ablation_iteration}    
\end{table}

\begin{table}
\centering
\resizebox{\linewidth}{!}{
  \begin{tabular}{l c c c c}
    \toprule
    Method & Params. & Memory& Training & FPS \\
    \midrule
    3DGS~\cite{kerbl3Dgaussians} & 94.1M & 5.3G & 29min & 171 \\
    SeaSplat~\cite{yang2025seasplat} & 29.4M & 5.1G & 126min & 169 \\
    WaterSplatting~\cite{li2025watersplatting} & 38.9M & 5.4G & 19min & 32 \\
    RainyScape~\cite{lyu2024rainyscape}  & 86.8M & 8.3G & 169min & 167 \\
    RobustSplat~\cite{2025RobustSplat}  & 81.0M & 8.9G & 29min & 168 \\
    WeatherGS~\cite{qian2025weathergs}   & 108.0M & 19.9G & 28min & 159 \\
    \midrule
    Ours & 78.4M & 9.3G & 77min & 162 \\
    \bottomrule
  \end{tabular}}
  \caption{Model complexity and runtime comparison under the H+R+S setting. Params., Memory, Training, and FPS denote parameter count, training memory usage, training time, and rendering speed, respectively.}
  \label{tab:runtime} 
  \end{table}
  
\subsection{Model Complexity and Runtime Analysis}
We compare the model complexity, training cost, and rendering efficiency of NimbusGS with several baselines in Tab.~\ref{tab:runtime}. Although NimbusGS introduces additional modules beyond the original 3DGS, its overall parameter count is lower, \ie, 78.4M versus 94.1M. Specifically, the CSM module contributes 2.1M parameters and the airlight MLP adds 0.2M, while the PLM component contains no learnable parameters. The remaining parameters are mainly associated with Gaussian primitives. Compared with 3DGS, NimbusGS requires fewer Gaussians because degradations are more effectively removed, reducing the need to compensate for them with additional primitives. In contrast, 3DGS tends to trigger additional densification to fit residual artifacts, leading to more Gaussians and a higher parameter count.

In terms of efficiency, NimbusGS uses 9.3G training memory, requires 77 minutes for training, and achieves 162 FPS rendering. These results indicate that the proposed degradation modeling introduces only moderate computational overhead while maintaining competitive training and rendering efficiency among weather-aware Gaussian splatting methods.

\begin{table}[t]
\centering
\setlength{\tabcolsep}{4pt}
\resizebox{\linewidth}{!}{
\begin{tabular}{l ccc ccc}
\toprule
\multirow{3}{*}{Method} &
\multicolumn{6}{c}{H+R} \\
\cmidrule(lr){2-7}
& \multicolumn{3}{c}{Stump} &
  \multicolumn{3}{c}{Wine} \\
\cmidrule(lr){2-4}\cmidrule(lr){5-7}
& PSNR$\uparrow$ & SSIM$\uparrow$ & LPIPS$\downarrow$
& PSNR$\uparrow$ & SSIM$\uparrow$ & LPIPS$\downarrow$ \\
\midrule
3DGS~\cite{kerbl3Dgaussians}
& 17.31 & 0.572 & 0.335
& 15.09 & 0.551 & 0.368 \\
SeaSplat~\cite{yang2025seasplat}
& 16.36 & 0.533 & 0.368
& 12.77 & 0.531 & 0.367 \\
WaterSplatting~\cite{li2025watersplatting}
& \second{17.49} & 0.538 & 0.340
& 14.45 & 0.518 & 0.305 \\
DerainNeRF~\cite{li2024derainnerf}
& 17.39 & 0.569 & 0.332
& 15.10 & 0.543 & 0.387\\
RainyScape~\cite{lyu2024rainyscape}
& 17.48 & 0.564 & \second{0.307}
& 15.51 & 0.600 & \second{0.228} \\
RobustSplat~\cite{2025RobustSplat}
& 17.29 & \second{0.586} & 0.344
& 15.53 & \second{0.629} & 0.315\\
WeatherGS~\cite{qian2025weathergs}
& 16.15 & 0.490 & 0.377
& 15.10 & 0.507 & 0.394 \\
OR~\cite{guo2024onerestore}+3DGS~\cite{kerbl3Dgaussians}
& 15.50 & 0.510 & 0.351
& \second{19.64} & 0.624 & 0.279 \\
\textbf{Ours}
& \first{21.33} & \first{0.648} & \first{0.277}
& \first{20.45} & \first{0.782} & \first{0.181} \\
\midrule
\multirow{3}{*}{Method} &
\multicolumn{6}{c}{H+S} \\
\cmidrule(lr){2-7}
& \multicolumn{3}{c}{Stump} &
  \multicolumn{3}{c}{Wine} \\
\cmidrule(lr){2-4}\cmidrule(lr){5-7}
& PSNR$\uparrow$ & SSIM$\uparrow$ & LPIPS$\downarrow$
& PSNR$\uparrow$ & SSIM$\uparrow$ & LPIPS$\downarrow$ \\
\midrule
3DGS~\cite{kerbl3Dgaussians}
& 16.82 & 0.575 & 0.322
& 15.78 & 0.642 & 0.256 \\
SeaSplat~\cite{yang2025seasplat}
& 16.08 & 0.498 & 0.417
& 16.15 & 0.617 & 0.264 \\
WaterSplatting~\cite{li2025watersplatting}
& 16.74 & 0.534 & 0.345
& 16.31 & 0.627 & 0.187 \\
DerainNeRF~\cite{li2024derainnerf}
& 16.00 & 0.499 & 0.401
& 16.03 & 0.630 & 0.294\\
RainyScape~\cite{lyu2024rainyscape}
& 17.02 & 0.564 & \second{0.309}
& 15.12 & 0.594 & 0.222  \\
RobustSplat~\cite{2025RobustSplat}
& 17.07 & \second{0.597} & 0.342
& 16.10 & 0.668 & 0.214 \\
WeatherGS~\cite{qian2025weathergs}
& \second{17.10} & 0.578 & 0.376
& 17.56 & 0.631 & 0.281 \\
OR~\cite{guo2024onerestore}+3DGS~\cite{kerbl3Dgaussians}
& 15.00 & 0.529 & 0.318
& \second{20.29} & \second{0.747} & \second{0.169} \\
\textbf{Ours}
& \first{20.27} & \first{0.657} & \first{0.257}
& \first{20.89} & \first{0.753} & \first{0.152}\\
\midrule
\multirow{3}{*}{Method} &
\multicolumn{6}{c}{R+S} \\
\cmidrule(lr){2-7}
& \multicolumn{3}{c}{Stump} &
  \multicolumn{3}{c}{Wine} \\
\cmidrule(lr){2-4}\cmidrule(lr){5-7}
& PSNR$\uparrow$ & SSIM$\uparrow$ & LPIPS$\downarrow$
& PSNR$\uparrow$ & SSIM$\uparrow$ & LPIPS$\downarrow$ \\
\midrule
3DGS~\cite{kerbl3Dgaussians}
& 19.87 & 0.579 & 0.348
& 21.59 & 0.698 & 0.292 \\
DerainNeRF~\cite{li2024derainnerf}
& 19.95 & 0.574 & 0.335
& 21.62 & 0.701 & 0.284\\
RainyScape~\cite{lyu2024rainyscape}
& 20.07 & 0.571 & \second{0.316}
& \second{22.35} & 0.738 & \first{0.123}\\
RobustSplat~\cite{2025RobustSplat}
& 19.90 & \second{0.588} & 0.329
& 21.58 & 0.702 & 0.289 \\
WeatherGS~\cite{qian2025weathergs}
& \second{21.12} & 0.523 & 0.348
& 21.41 & 0.674 & 0.297 \\
OR~\cite{guo2024onerestore}+3DGS~\cite{kerbl3Dgaussians}
& 15.17 & 0.536 & 0.334
& 22.33 & \second{0.756} & 0.229 \\
\textbf{Ours}
& \first{21.63} & \first{0.600} & \first{0.303}
& \first{23.27} & \first{0.811} & \second{0.142} \\
\bottomrule
\end{tabular}}\vspace{-2mm}
\caption{Quantitative comparisons on the additional Stump and Wine scenes. The three blocks correspond to the H+R, H+S, and R+S settings. The \colorbox{red!25}{best} and \colorbox{orange!25}{second-best} scores are color-encoded for clarity.}\vspace{-4mm}
\label{tab:stump_wine}
\end{table}

\begin{table*}[t]
\centering
\setlength{\tabcolsep}{3.5pt}
\resizebox{\linewidth}{!}{
\begin{tabular}{l ccc ccc ccc ccc ccc}
\toprule
\multirow{2}{*}{Method} &
\multicolumn{3}{c}{Bicycle} &
\multicolumn{3}{c}{Garden} &
\multicolumn{3}{c}{Factory} &
\multicolumn{3}{c}{Tanabata} &
\multicolumn{3}{c}{Average} \\
\cmidrule(lr){2-4}\cmidrule(lr){5-7}\cmidrule(lr){8-10}\cmidrule(lr){11-13}\cmidrule(lr){14-16}
& PSNR$\uparrow$ & SSIM$\uparrow$ & LPIPS$\downarrow$
& PSNR$\uparrow$ & SSIM$\uparrow$ & LPIPS$\downarrow$
& PSNR$\uparrow$ & SSIM$\uparrow$ & LPIPS$\downarrow$
& PSNR$\uparrow$ & SSIM$\uparrow$ & LPIPS$\downarrow$
& PSNR$\uparrow$ & SSIM$\uparrow$ & LPIPS$\downarrow$ \\
\midrule
3DGS~\cite{kerbl3Dgaussians}
& 15.24 & 0.491 & 0.370
& 15.95 & 0.530 & 0.376
& 13.90 & 0.534 & 0.406
& 13.58 & 0.445 & 0.428
& 14.66 & 0.500 & 0.395 \\
SeaSplat~\cite{yang2025seasplat}
& 15.48 & 0.523 & 0.354
& 13.56 & 0.558 & 0.359
& 14.64 & 0.621 & 0.329
& 14.58 & 0.564 & 0.324
& 14.56 & 0.566 & 0.341 \\
WaterSplatting~\cite{li2025watersplatting}
& 16.69 & 0.537 & 0.360 
& 17.42 & 0.592 & 0.312 
& 14.71 & 0.602 & 0.307 
& 14.45 & 0.518 & 0.309 
& 15.81 & 0.562 & 0.322 \\
DerainNeRF~\cite{li2024derainnerf}
& 16.10 & 0.499 & 0.401 
& 16.27 & 0.556 & 0.350 
& 14.29 & 0.583 & 0.372 
& 14.31 & 0.503 & 0.407 
& 15.24 & 0.535 & 0.382 \\
RainyScape~\cite{lyu2024rainyscape}
& 16.66 & 0.504 & 0.387
& 18.22 & \second{0.653} & \second{0.267}
& 14.75 & 0.642 & \second{0.268}
& 14.65 & 0.527 & \second{0.306} 
& 16.07 & 0.581 & \second{0.307} \\
RobustSplat~\cite{2025RobustSplat}
& 16.28 & 0.498 & 0.379
& 17.43 & 0.603 & 0.322
& 13.97 & 0.542 & 0.404
& 13.63 & 0.443 & 0.429
& 15.32 & 0.521 & 0.383 \\
WeatherGS~\cite{qian2025weathergs}
& 15.94 & 0.479 & 0.414 
& 17.56 & 0.509 & 0.436 
& 14.62 & 0.569 & 0.395 
& 14.60 & 0.533 & 0.405 
& 15.68 & 0.522 & 0.412  \\
OR~\cite{guo2024onerestore}+3DGS~\cite{kerbl3Dgaussians}
& \second{17.01} & \second{0.552} & \second{0.353}
& \second{20.28} & 0.636 & 0.311
& \second{18.43} & \second{0.698} & 0.300
& \second{20.11} & \second{0.637} & 0.338
& \second{18.95} & \second{0.630} & 0.325 \\
\textbf{Ours}
& \first{19.71} & \first{0.651} & \first{0.321}
& \first{22.96} & \first{0.727} & \first{0.200}
& \first{21.87} & \first{0.836} & \first{0.151}
& \first{22.17} & \first{0.722} & \first{0.229}
& \first{21.67} & \first{0.734} & \first{0.225} \\
\bottomrule
\end{tabular}}
\caption{Quantitative comparisons on H+R scenes. The \colorbox{red!25}{best} and \colorbox{orange!25}{second-best} scores are color-encoded for clarity.}
\label{tab:h+r}
\end{table*}

\begin{table*}[t]
\centering
\setlength{\tabcolsep}{3.5pt}
\resizebox{\linewidth}{!}{
\begin{tabular}{l ccc ccc ccc ccc ccc}
\toprule
\multirow{2}{*}{Method} &
\multicolumn{3}{c}{Bicycle} &
\multicolumn{3}{c}{Garden} &
\multicolumn{3}{c}{Factory} &
\multicolumn{3}{c}{Tanabata} &
\multicolumn{3}{c}{Average} \\
\cmidrule(lr){2-4}\cmidrule(lr){5-7}\cmidrule(lr){8-10}\cmidrule(lr){11-13}\cmidrule(lr){14-16}
& PSNR$\uparrow$ & SSIM$\uparrow$ & LPIPS$\downarrow$
& PSNR$\uparrow$ & SSIM$\uparrow$ & LPIPS$\downarrow$
& PSNR$\uparrow$ & SSIM$\uparrow$ & LPIPS$\downarrow$
& PSNR$\uparrow$ & SSIM$\uparrow$ & LPIPS$\downarrow$
& PSNR$\uparrow$ & SSIM$\uparrow$ & LPIPS$\downarrow$ \\
\midrule
3DGS~\cite{kerbl3Dgaussians}
& 17.19 & 0.554 & 0.327 
& 14.21 & 0.507 & 0.411
& 13.87 & 0.601 & 0.282
& 15.16 & 0.615 & 0.261
& 15.10 & 0.569 & 0.320 \\
SeaSplat~\cite{yang2025seasplat}
& 15.73 & 0.525 & 0.348 
& 14.66 & 0.513 & 0.400 
& 13.98 & 0.626 & 0.247 
& 15.18 & 0.625 & 0.339
& 14.88 & 0.572 & 0.333 \\
WaterSplatting~\cite{li2025watersplatting}
& 16.47 & 0.530 & 0.325 
& 14.10 & 0.512 & 0.422 
& 14.42 & 0.613 & 0.276 
& 15.37 & 0.623 & 0.251 
& 15.09 & 0.569 & 0.318 \\
DerainNeRF~\cite{li2024derainnerf}
& 17.12 & 0.558 & 0.331 
& 14.32 & 0.576 & 0.378 
& 14.12 & 0.611 & 0.279 
& 15.16 & 0.617 & 0.288 
& 15.18 & 0.590 & 0.319 \\
RainyScape~\cite{lyu2024rainyscape}
& 17.33 & 0.567 & \second{0.256} 
& 14.52 & 0.600 & 0.369 
& 14.08 & 0.631 & 0.241 
& 15.75 & 0.641 & 0.211
& 15.42 & 0.609 & 0.269 \\
RobustSplat~\cite{2025RobustSplat}
& 17.41 & 0.556 & 0.323
& 14.17 & 0.552 & 0.416
& 13.86 & 0.600 & 0.275
& 15.34 & 0.637 & 0.224
& 15.19 & 0.586 & 0.309 \\
WeatherGS~\cite{qian2025weathergs}
& \second{17.92} & \second{0.579} & 0.312 
& 14.34 & 0.566 & 0.419 
& 13.82 & 0.571 & 0.407 
& 15.57 & 0.536 & 0.339 
& 15.41 & 0.563 & 0.369 \\
OR~\cite{guo2024onerestore}+3DGS~\cite{kerbl3Dgaussians}
& 16.24 & 0.475 & 0.396
& \second{23.09} & \second{0.693} & \second{0.295}
& \second{21.34} & \second{0.818} & \second{0.178}
& \second{23.25} & \second{0.801} & \second{0.159}
& \second{20.98} & \second{0.696} & \second{0.257} \\
\textbf{Ours}
& \first{19.11} & \first{0.625} & \first{0.251}
& \first{23.15} & \first{0.709} & \first{0.232}
& \first{21.48} & \first{0.839} & \first{0.136}
& \first{23.50} & \first{0.809} & \first{0.116}
& \first{21.81} & \first{0.745} & \first{0.183} \\
\bottomrule
\end{tabular}}
\caption{Quantitative comparisons on H+S scenes. The \colorbox{red!25}{best} and \colorbox{orange!25}{second-best} scores are color-encoded for clarity.}
\label{tab:h+s}
\end{table*}

\begin{table*}[!t]
\centering
\setlength{\tabcolsep}{3.5pt}
\resizebox{\linewidth}{!}{
\begin{tabular}{l ccc ccc ccc ccc ccc}
\toprule
\multirow{2}{*}{Method} &
\multicolumn{3}{c}{Bicycle} &
\multicolumn{3}{c}{Garden} &
\multicolumn{3}{c}{Factory} &
\multicolumn{3}{c}{Tanabata} &
\multicolumn{3}{c}{Average} \\
\cmidrule(lr){2-4}\cmidrule(lr){5-7}\cmidrule(lr){8-10}\cmidrule(lr){11-13}\cmidrule(lr){14-16}
& PSNR$\uparrow$ & SSIM$\uparrow$ & LPIPS$\downarrow$
& PSNR$\uparrow$ & SSIM$\uparrow$ & LPIPS$\downarrow$
& PSNR$\uparrow$ & SSIM$\uparrow$ & LPIPS$\downarrow$
& PSNR$\uparrow$ & SSIM$\uparrow$ & LPIPS$\downarrow$
& PSNR$\uparrow$ & SSIM$\uparrow$ & LPIPS$\downarrow$ \\
\midrule
3DGS~\cite{kerbl3Dgaussians}
& 19.25 & 0.590 & 0.332 
& 23.39 & 0.714 & 0.245
& 18.29 & 0.685 & 0.306
& 22.79 & 0.765 & 0.241
& 20.93 & 0.688 & 0.281 \\
DerainNeRF~\cite{li2024derainnerf}
& 18.37 & 0.533 & 0.369 
& 23.12 & 0.712 & 0.295 
& 18.62 & 0.675 & 0.297 
& 22.89 & 0.762 & 0.255 
& 20.75 & 0.670 & 0.304 \\
RainyScape~\cite{lyu2024rainyscape}
& \second{19.57} & \second{0.614} & \second{0.285}
& \second{24.25} & \second{0.729} & \second{0.216}
& 22.12 & \second{0.805} & \second{0.127}
& 24.07 & \second{0.808} & \second{0.117} 
& \second{22.50} & \second{0.739} & \second{0.186} \\
RobustSplat~\cite{2025RobustSplat}
& 19.56 & 0.588 & 0.348 
& 23.70 & 0.718 & 0.271
& 18.99 & 0.701 & 0.295
& 22.86 & 0.785 & 0.244
& 21.27 & 0.698 & 0.289 \\
WeatherGS~\cite{qian2025weathergs}
& 17.47 & 0.539 & 0.413 
& 20.71 & 0.618 & 0.442 
& 19.11 & 0.562 & 0.364 
& 22.01 & 0.733 & 0.243 
& 19.82 & 0.613 & 0.365 \\
OR~\cite{guo2024onerestore}+3DGS~\cite{kerbl3Dgaussians}
& 16.91 & 0.533 & 0.368
& 24.02 & 0.713 & 0.252
& \second{22.13} & 0.757 & 0.219
& \second{24.35} & 0.814 & 0.177
& 21.85 & 0.704 & 0.254 \\
\textbf{Ours}
& \first{20.03} & \first{0.644} & \first{0.278}
& \first{24.81} & \first{0.751} & \first{0.194}
& \first{23.82} & \first{0.859} & \first{0.120}
& \first{24.81} & \first{0.845} & \first{0.107}
& \first{23.36} & \first{0.774} & \first{0.174} \\
\bottomrule
\end{tabular}}
\caption{Quantitative comparisons on R+S scenes. The \colorbox{red!25}{best} and \colorbox{orange!25}{second-best} scores are color-encoded for clarity.}
\label{tab:r+s}
\end{table*}

\section{Additional Weather-Effects Evaluation}
This section extends the hybrid-weather evaluation beyond the configurations considered in the main paper. For clarity, the main paper adopts a task-aligned comparison protocol. Single-weather methods are evaluated only on the corresponding single-weather tasks, and methods designed for hybrid-weather scenarios are evaluated on hybrid-weather settings. This ensures a fair and meaningful comparison within the scope of each task. In the supplementary material, we relax this restriction and include all baselines that are able to generate images under the corresponding degradations. This broader comparison allows a more comprehensive examination of model behavior under mixed adverse conditions.

We include three additional hybrid compositions, H+R, H+S, and R+S, to more thoroughly stress-test NimbusGS under mixed adverse conditions that combine volumetric degradations with particle-based artifacts. To further enrich scene diversity under these challenging settings, we also incorporate two additional scenes used exclusively for the supplementary experiments, broadening the spatial layouts and weather interactions while keeping the training and evaluation protocol identical to that in the main benchmark.

\subsection{Quantitative Results}
We first report quantitative results on the four main scenes (Bicycle, Garden, Factory, and Tanabata) under the H+R, H+S, and R+S configurations in Tab.~\ref{tab:h+r}, Tab.~\ref{tab:h+s}, and Tab.~\ref{tab:r+s}. Across all three hybrid-weather settings, NimbusGS consistently achieves the best performance in terms of PSNR, SSIM, and LPIPS. In contrast, directly applying 3DGS or recent weather-aware Gaussian and NeRF variants leads to a substantial quality drop once heterogeneous haze–rain–snow interactions are present. The single-image baseline OR~\cite{guo2024onerestore} combined with 3DGS (\ie, OR+3DGS) forms a strong competitor and typically ranks second, but still falls short of NimbusGS by a notable margin, particularly on challenging scenes such as Bicycle and Tanabata where both structural consistency and perceptual quality are severely affected by mixed degradations.

To further assess generalization under different spatial layouts, we additionally evaluate two new scenes, Stump from Mip-NeRF360~\cite{barron2022mip} and Wine from Deblur-NeRF~\cite{ma2022deblur}, as shown in Tab.~\ref{tab:stump_wine}. The performance trends match those observed on the main benchmark: NimbusGS maintains the top results across all hybrid-weather combinations, while OR+3DGS remains competitive but continues to underperform our method. These supplementary experiments confirm that NimbusGS scales reliably to diverse scene structures and more complex mixtures of weather degradations.

\subsection{Additional Visual Comparisons}
\begin{figure*}[!t]
	\centering
	\includegraphics[width=0.98\linewidth]{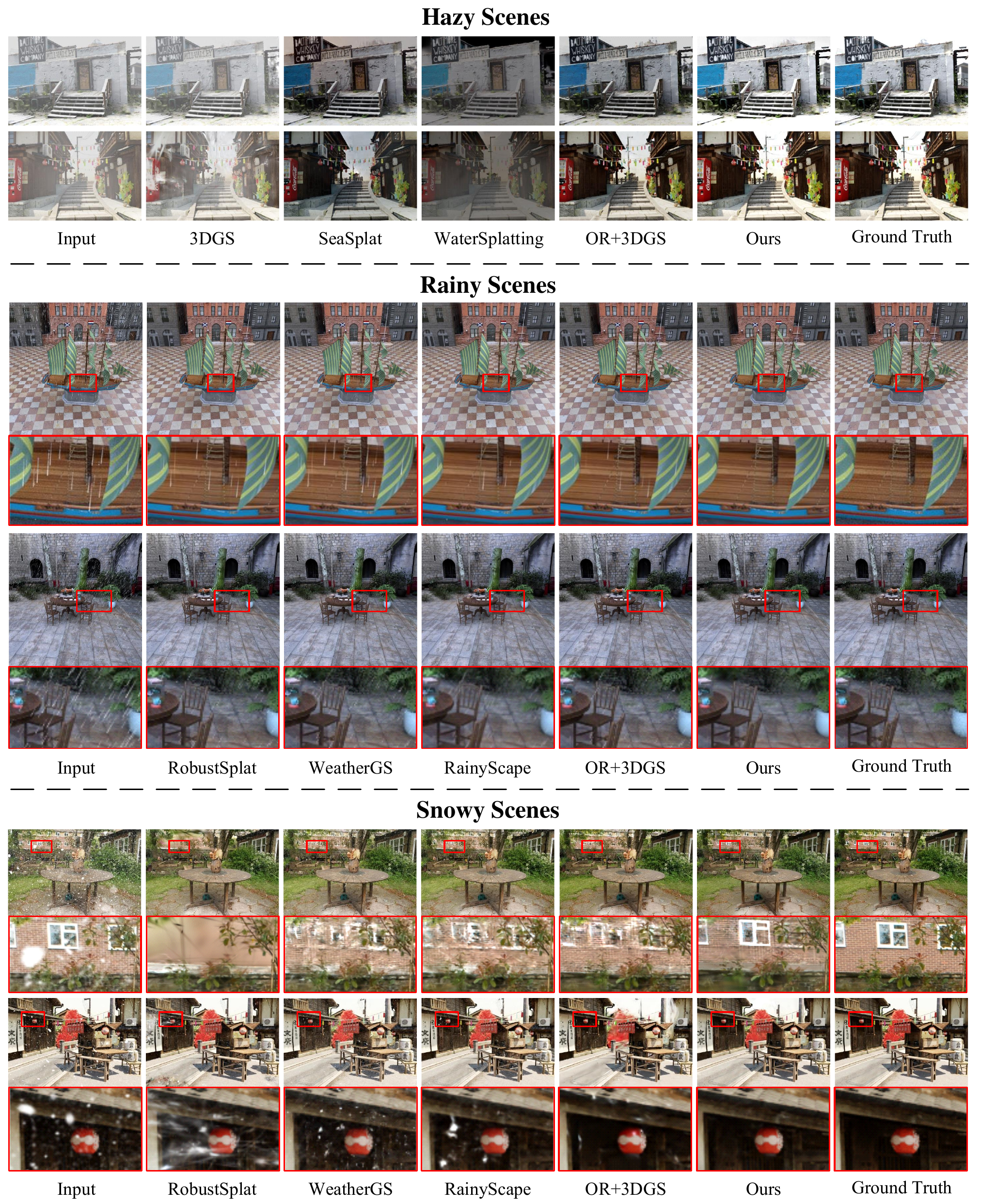}
	\caption{Qualitative comparisons on single-weather scenes (haze, rain, snow).}
	\label{fig:single}
\end{figure*}

\begin{figure*}[!t]
	\centering
	\includegraphics[width=0.98\linewidth]{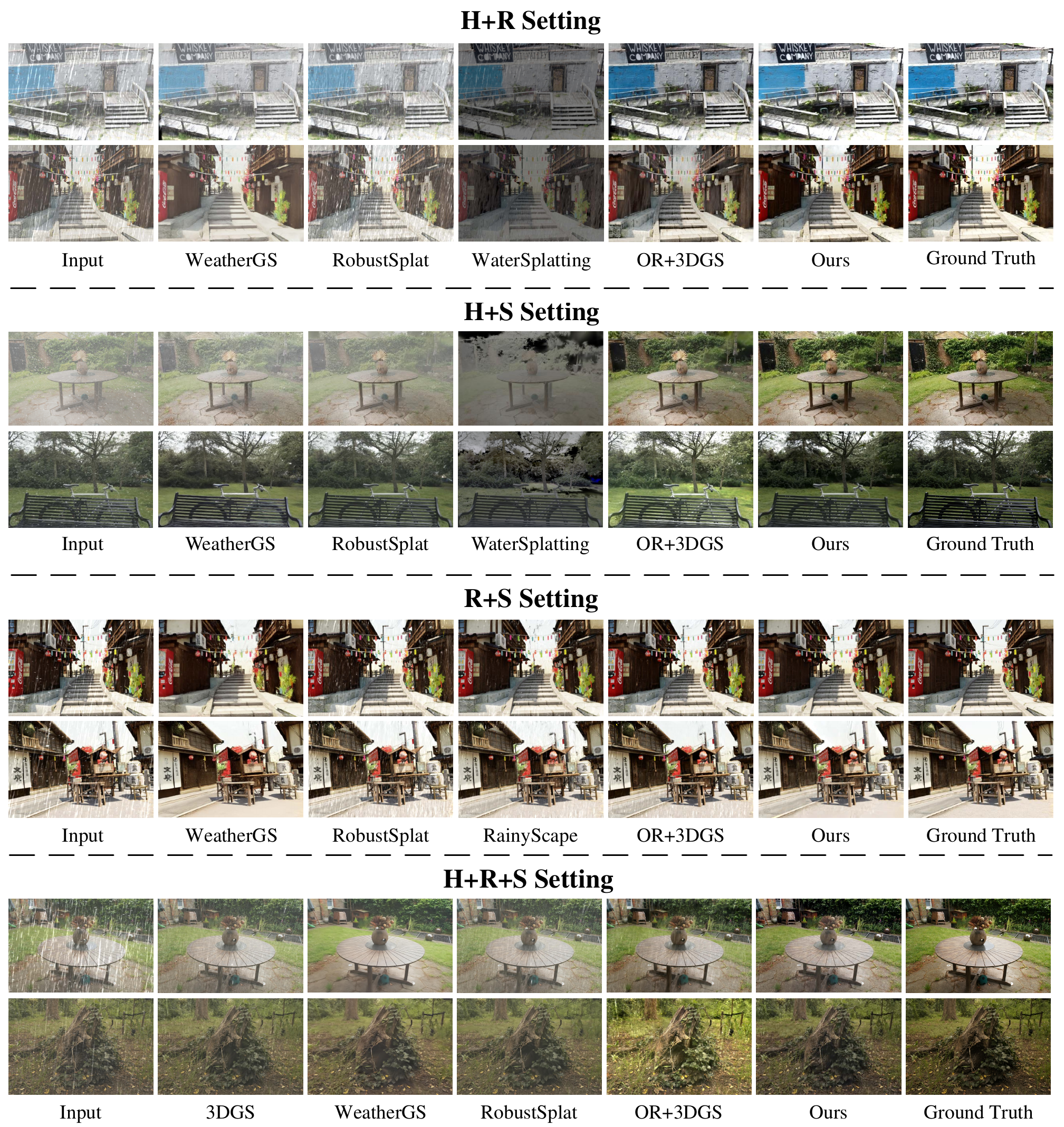}
	\caption{Qualitative comparisons on hybrid-weather scenes (H+R, H+S, R+S, H+R+S). Best viewed zoomed in.}
	\label{fig:mix}
\end{figure*}

We provide additional qualitative comparisons for all seven weather settings, covering both the single-weather tasks and the hybrid-weather tasks. Figs.~\ref{fig:single} and~\ref{fig:mix} show supplementary visual examples for these two groups of settings. Due to space limitations, each figure includes the top-performing baselines for the corresponding task.

Specifically, Fig.~\ref{fig:single} presents the single-weather comparisons, where NimbusGS preserves geometry and appearance quality across haze-, rain-, and snow-induced degradations. Fig.~\ref{fig:mix} shows the results for the four hybrid-weather settings that combine volumetric scattering and particle-based artifacts, demonstrating the effectiveness of NimbusGS under more complex adverse weather conditions.

\section{Limitation}
Despite its effectiveness under adverse weather conditions, NimbusGS may drop in sparse-view settings. This is because removing particulate residuals reveals regions without appearance cues, while the model lacks priors to recover them. These regions then rely on cross-view information, often insufficient with few views, which remains an open challenge for future work.

{
    \small
    \bibliographystyle{ieeenat_fullname}
    \bibliography{main}
}

{
    \small
    \bibliographystyle{ieeenat_fullname}
    \bibliography{main}
}

\end{document}